\documentclass[acmsmall,screen,
]{acmart}

\renewcommand\footnotetextcopyrightpermission[1]{}
\usepackage{lipsum}
\usepackage{algorithm}
\usepackage{algorithmic}

\usepackage{booktabs}
\usepackage{multirow}
\usepackage{makecell}

\AtBeginDocument{%
  \providecommand\BibTeX{{%
    \normalfont B\kern-0.5em{\scshape i\kern-0.25em b}\kern-0.8em\TeX}}}

\begin{document}

\title{EF-Train: Enable Efficient On-device CNN Training on FPGA Through Data Reshaping for Online Adaptation or Personalization}

\author{Yue Tang}
\affiliation{%
  \institution{University of Pittsburgh}
  \city{Pittsburgh}
  \country{USA}}
\email{yut51@pitt.edu}

\author{Xinyi Zhang}
\affiliation{%
  \institution{University of Pittsburgh}
  \city{Pittsburgh}
  \country{USA}}
\email{xinyizhang@pitt.edu}

\author{Peipei Zhou}
\affiliation{%
  \institution{University of Pittsburgh}
  \city{Pittsburgh}
  \country{USA}}
\email{peipei.zhou@pitt.edu}

\author{Jingtong Hu}
\affiliation{%
  \institution{University of Pittsburgh}
  \city{Pittsburgh}
  \country{USA}}
\email{jthu@pitt.edu}







\renewcommand{\shortauthors}{Tang and Hu, et al.}

\begin{abstract}
  Conventionally, DNN models are trained once in the cloud and deployed in edge devices such as cars, robots, or unmanned aerial vehicles (UAVs) for real-time inference. However, there are many cases that require the models to adapt to new environments, domains, or new users. In order to realize such domain adaption or personalization, the models on devices need to be continuously trained on the device. In this work, we design EF-Train, an efficient DNN training accelerator with a unified channel-level parallelism-based convolution kernel that can achieve end-to-end training on resource-limited low-power edge-level FPGAs. It is challenging to implement on-device training on resource-limited FPGAs due to the low efficiency caused by different memory access patterns among forward, backward propagation, and weight update. Therefore, we developed a data reshaping approach with intra-tile continuous memory allocation and weight reuse. An analytical model is established to automatically schedule computation and memory resources to achieve high energy efficiency on edge FPGAs. The experimental results show that our design achieves 46.99 GFLOPS and 6.09 GFLOPS/W in terms of throughput and energy efficiency, respectively.
\end{abstract}






\maketitle

\section{Introduction}\label{sec:introduction}
Deep Neural Networks (DNNs) have been widely used in edge devices such as cars, robotics~\cite{granter2017alphago}, and unmanned aerial vehicles (UAVs)~\cite{zeng2020federated}, to accomplish various tasks, including autonomous driving, object detection, etc. FPGAs are promising platforms with higher computational density, communication bandwidth, and energy efficiency, and can be configured based on different tasks. Nowadays, FPGAs have been widely used in various edge device domains. For example, edge-scale FPGAs are commonly utilized in object detection tasks with high frames per second and low power consumption~\cite{hao2019fpga}. The Corazon-AI built on Xilinx Zynq is a perfect fit for various computer-vision-based applications including video surveillance, advanced driver-assistance systems (ADAS), medical robotics, industrial automation, and augmented reality~\cite{Corazon}. Combined with reconfigurability, FPGAs have been adopted in several autonomous platforms such as pony.ai~\cite{ponyai} and ZF ProAI~\cite{ZFProAI}. In medical applications, an FPGA-based low-latency multi-layer perception (MLP) processor for real-time cancer detection has been developed, since FPGA-based design can directly interface with sensors, display devices, and reduce data movement delays~\cite{sanaullah2018real}. Burger et al. applied an FPGA-based embedded device to monitor users’ electrocardiograms (ECGs) in a pervasive internet-of-things (IoT) system~\cite{burger2020embedded}.  FPGAs have also been well performed in other areas such as agricultural robots~\cite{lammie2019low}, UAVs~\cite{zhang2019skynet}, etc. 

In traditional FPGA-based edge device applications, DNNs are pre-trained in the cloud before being deployed in FPGAs, which is not efficient for domain adaption. When the environments, tasks, or users change, data needs to be collected from the edge FPGAs and transmitted to the cloud. Then, the cloud retrains a new model, transmitting the model back to the edge devices. The whole process is inefficient and time-consuming. Therefore, it is often desirable for edge FPGAs to continuously and locally learn from new data. Such on-device learning can directly improve model accuracy and adapt to new environments. {
Currently, several algorithms have been proposed to enable edge devices to achieve domain adaption locally. For example, a MobileDA framework~\cite{yang2020mobileda} has been developed to allow a novel teacher network trained in the server to distill the knowledge for a student network running in the edge device, and the algorithm was employed on an embedded GPU and NVIDIA Jetson TX2. A transductive transfer learning model HDCNN~\cite{khan2018scaling} has been proposed to allow adaptation without requiring collecting large volumes of labeled training data in the target domain, and the algorithm was tested on 1080 Ti GPU. To implement these complex and fantastic software-level algorithms on FPGA-based edge devices, an FPGA-based training accelerator is indispensable. However, traditional FPGA-based edge device applications lack such hardware-level designs for training operations, which prevents FPGA-based devices from applying these algorithms directly.}

Furthermore, directly training Convolutional Neural Network (CNN) models on local FPGAs can facilitate personalization. For example, in some medical applications such as home monitoring~\cite{mendoncca2021method}, long-term ECG monitoring~\cite{rana2020comparison}, etc., the distinction of different users' physical conditions will impact data distribution, so models need to be fine-tuned based on specific users. The system in~\cite{burger2020embedded} utilized cloud services to log a user’s condition over time and continuously improve the system’s performance. It would be more effective if models could be directly updated on the FPGA device. Besides, learning at the edge can provide better privacy since users do not need to upload data into the central cloud~\cite{xu2018deeptype}.

However, it has been challenging to implement on-device training on FPGAs. Previous works mainly focused on implementing CNN inference on FPGAs. For example, Zhang et al.~\cite{JasonOptimizing} exploited various optimization techniques including loop unrolling, loop tiling, and loop transformation on the FPGA accelerator, and proposed a roofline model to quantitatively analyze its computing throughput and required memory bandwidth. 
Various designs~\cite{DNNbuilder,JasonOvercoming} have been proposed to map well-trained neural networks on FPGAs for inference with high throughput and low latency. Compared with CNN inference, it is more complex to efficiently implement CNN training on FPGAs in terms of the following aspects. 
First, the inference process only includes forward propagation (FP), whereas the training process includes FP, backward propagation (BP), and weight update (WU), which leads to a 3X computation operation count with more types of operations~\cite{choi2018trainware}.
Second, the large volume of activation data in FP needs to be used in BP and WU, and the loss data generated in BP is also required in WU. Such data dependency across multiple layers makes it difficult for on-board memory management and data reusing in dynamic random access memory (DRAM) in an end-to-end training system~\cite{tao2020challenges}. 
Third, since FP, BP, and WU have different memory access patterns, simply using the memory optimizations of FP in the whole training process leads to low memory access efficiency in BP and WU. Because of the above-mentioned challenges, CNN training on FPGAs has not been comprehensively investigated.

Recently, several FPGA-based architectures have been designed to accelerate training on large scale FPGAs. F-CNN~\cite{zhao2016f} first performed FP and BP on a Maxeler MPC-X dataflow FPGA node but WU on CPU. Designs such as~\cite{venkataramanaiah2019automatic,luo2020towards} aimed to further reduce the CNN training latency and improve throughput. However, these works mainly focused on cloud-level devices with abundant resources. A straightforward training implementation on edge FPGAs is still challenging.

\begin{figure}[h]
  \centering
  \includegraphics[width=4.6in
  ]{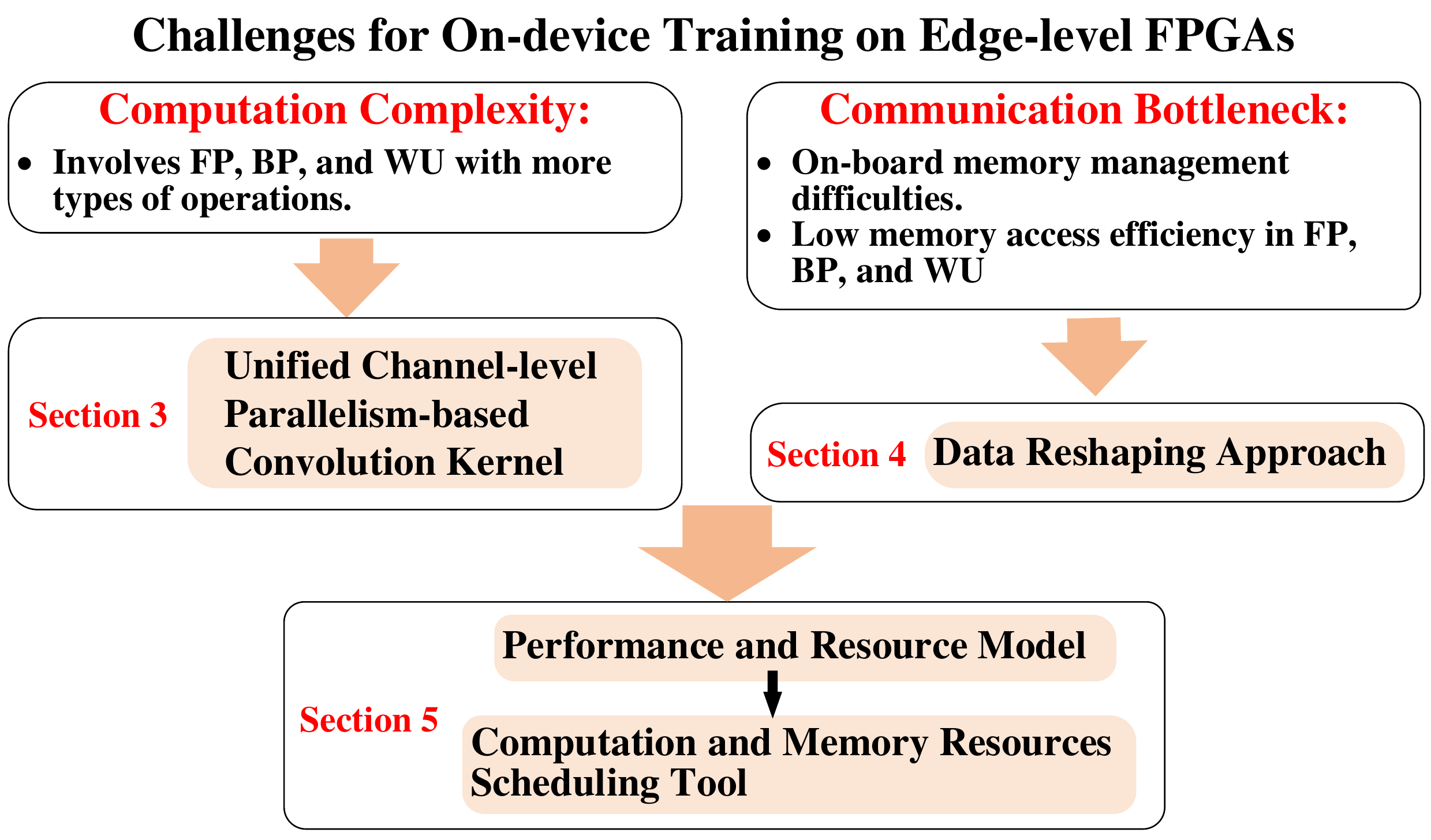}
  \caption{Overview of our design framework.}
  \label{fig:overview}
  \Description{fig:overview}
\end{figure}

To tackle the challenges in implementing on-device training on edge-level FPGAs, we propose EF-Train, a new efficient FPGA-based training accelerator with a unified channel-level parallelism-based convolution kernel to handle the computation complexity. {
The unified kernel means that it processes convolution operations of FP, BP, and WU in the training phase utilizing the same computation resources on the FPGA chip. The channel-level parallelism means that the kernel allocates these computation resources to process multiple channels of feature maps in parallel.} We also propose a data reshaping approach to solve the communication bottleneck in realistic end-to-end training processes. The overview of the design framework is shown in~Fig.~\ref{fig:overview}. {
The data reshaping approach is a compile-time optimization that achieves intra-tile and inter-tile memory access continuity and weight reuse in mini-batch training.} The proposed design can be implemented on resource-limited FPGAs without sacrificing precision. Neural networks can be trained on both small batches and large batches. {
Since training and inference are conducted separately in realistic applications, and FPGAs are configurable to implement different designs on the same hardware platform for different applications, our design can be applied to those well-developed FPGA-based inference devices. In a relatively long life cycle of the inference phase, the original design can guarantee high throughput and low latency. If users or the environments change, the device can be switched to implement our design immediately to learn from local data for online adaptation or personalization rather than transmitting data to the cloud centers and waiting for the cloud centers to transmit the well-trained model back to the device.}  Our main contributions are as follows.
\begin{itemize}
\item We propose EF-Train, an efficient FPGA-based CNN training accelerator with a unified convolution kernel to process FP, BP, and WU with full precision. The accelerator exploits channel-level parallelism to achieve high computation utilization for both small and large batch sizes. {
Our accelerator supports end-to-end CNN training with convolutional (Conv) layers, fully  connected (FC) layers, batch normalization (BN) layers, rectified linear unit (ReLU) layers, and pooling layers (Section~\ref{sec:accelerator}).}
\item We propose a data reshaping approach to solve the off-chip communication bottleneck. The features and weights are stored in off-chip memory with intra-tile continuous memory allocation to remove discontinuous memory accesses within a tile. We also reduce inter-tile discontinuous memory accesses by scheduling loop orders between tiles. We further exploit weight reuse among multiple images in a mini-batch to improve communication efficiency when the batch size is larger than one (Section~\ref{sec:memory}).
\item We build a performance and resource model for the proposed accelerator. Based on the model, a computation and memory resources scheduling tool is established to determine design parameters for different FPGA devices and different neural networks (Section~\ref{sec:model}).
\item We deploy the training process on PYNQ-Z1 and ZCU102 for various CNNs on both Cifar-10 and ImageNet datasets. Experimental results show
that our design can achieve 46.99 GFLOPS and 6.09 GFLOPS/W in terms of throughput and energy efficiency, respectively (Section~\ref{sec:experiment}).
\end{itemize}

\section{Background and Motivations}
\label{sec:Background}

\subsection{CNN Training}
\label{sec:CNNs Training}

\begin{figure}[h]
  \centering
  \includegraphics[width=5.4in
  ]{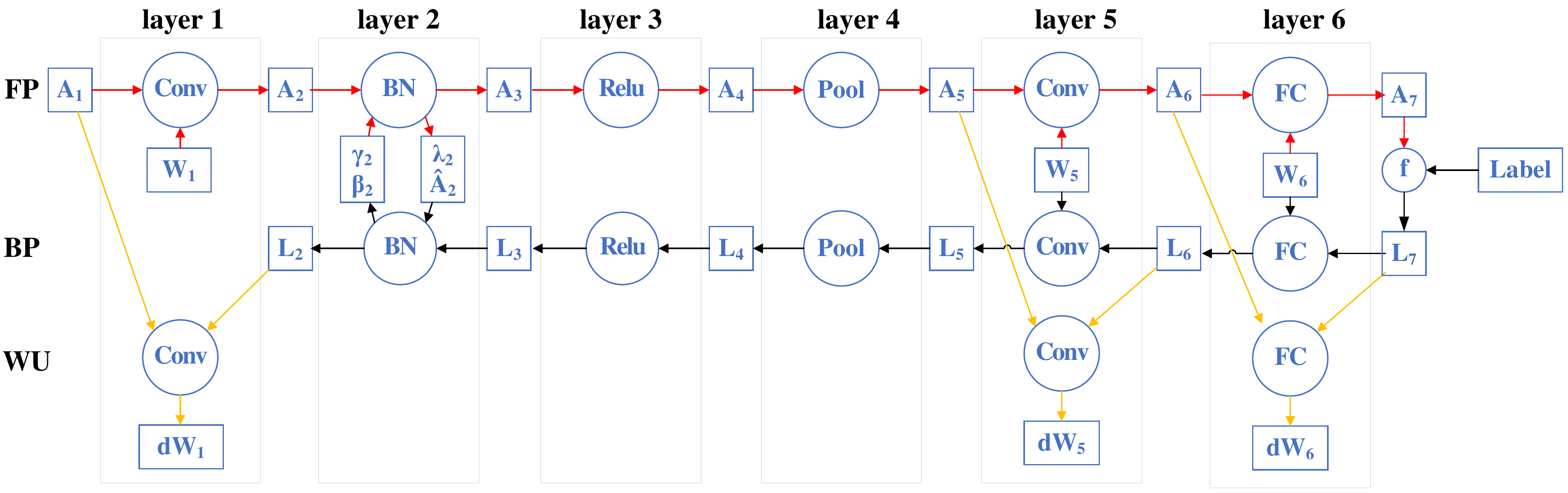}
  \caption{
  CNN training process.}
  \label{fig:training process}
  \Description{fig:training process}
\end{figure}

The training process of a five-layer CNN is shown in Fig.~\ref{fig:training process}, including the FP process (red arrows), the BP process (black arrows), and the WU process (yellow arrows). {
The network includes two Conv layers, one FC layer, one BN layer, one ReLU layer, and one pooling layer which are practical and can make up most neural networks in real-world scenarios.}

In the FP process, activation is propagated layer by layer. In a Conv layer such as layer 1, the input activation $A_1$ conducts multiply-accumulate
(MAC) operations with the weights $W_1$. {
A BN layer is always followed by a Conv layer. In layer 2, the inputs of the BN layer include the input activation $A_2$ and learnable parameters $\gamma_2$ and $\beta_2$. The immediate outputs include $\lambda_2$, $\hat A_2$, and output activation $A_3$. $A_3$ then goes through the ReLU and pooling layers.} Finally, the FC layer provides classification results for the input image.

In the BP process, the loss will be calculated and propagated back to the first layer. The loss of the last layer is calculated by the loss function $f$. This paper adopts the most commonly used cross-entropy loss function. The stochastic gradient descent (SGD) is applied in CNN training. {
In a Conv layer, layer 5 for example, the loss $L_6$ needs to be padded first to ensure the size of the convolution results $L_5$ is the same as the size of $A_5$. The tensors for weights $W_5$ are transposed on dimensions for output channels and input channels. The original kernel tensors need to be flipped. Then, $L_5$ is calculated by the convolution operation between the transposed weights and $L_6$. In the maximum pooling layer, layer 4, $A_4$ is compared with $A_5$ to determine which element on $L_4$ should obtain the value from the corresponding position on $L_5$. If layer 4 is an average pooling layer, the values for each element of each patch in $L_5$ will be directly accumulated and propagated to $L_4$. In the ReLU layer, layer 3, an element of $L_3$ will return zero if the value in the same position of $A_3$ is less than zero. Otherwise, the value of the corresponding position in $L_4$ will be propagated back. In the BN layer, layer 2, $\gamma_2$ and $\beta_2$ are updated according to the value of $\lambda_2$, $\hat A_2$, and $L_3$. Then the loss is propagated back to $L_2$.}

The gradients of weights in layer $i$ is calculated after the loss of layer $i+1$ is propagated. In layer 5, $dW_5$ is calculated by conducting MAC operations for $A_5$ and $L_6$. The gradients are accumulated inside a batch. In a mini-batch, after the above-mentioned operations are conducted for all images, $W_5$ will be updated by subtracting $dW_5\times learning\ rate$.

\subsection{Related Works}
\label{sec:Related works}

{
\noindent\textbf{DRAM Access Issues for Current FPGA-based Inference Accelerators:} Currently, FPGAs have been widely adopted in edge domains thanks to the well-developed FPGA-based inference accelerators~\cite{JasonOptimizing,DNNbuilder}. Among the inference accelerators, many works~\cite{JasonOptimizing,jiang2019achieving,kao2020confuciux} mainly focused on selecting optimal design parameters to improve the acceleration performance for individual Conv layers. Optimizing techniques such as loop tiling, loop unrolling are adopted by these works. Although the proposed algorithms achieved higher performance efficiency on a given Conv layer, the proposed designs only presented isolated accelerators without completing end-to-end inference where all layers of a neural network are tested continuously. In an end-to-end system, the layers' intermediate results are usually transferred between on-chip buffer and off-chip DRAM due to the limited on-chip storage size, so the impact of off-chip memory accesses should be considered in realistic scenarios. For most edge-level FPGAs, direct memory access (DMA) is a commonly used effective data swapping way for continuous address data reading. {
However, in current FPGA-based DNN deployments, when the on-chip memory cannot hold all the features and weights of a Conv layer, data need to be fetched and processed in tiles based on the computation pattern. Such tiling schemes break the continuity of data addresses in DRAM and thus reduce the DMA transmission efficiency. The detailed analysis will be further discussed in Section~\ref{sec:memory}.} This discontinuity can degrade the DMA transferring speed from about 8GB/s to around 1GB/s~\cite{guan2017fp}. The optimal algorithms proposed in the above-mentioned accelerators are based on the assumption that data are well pre-allocated between adjacent layers so tiles can be loaded from and stored back to the DRAM continuously. However, in actual end-to-end systems, such allocation overhead is extremely large compared to the acceleration time. 

\noindent\textbf{Solutions for The DRAM Access Issues in The Inference Phase:} Issues related to DRAM memory access have been addressed in recent works. For example, ROMANet~\cite{putra2021romanet} proposed a design space exploration (DSE) by searching for the appropriate data partitioning and scheduling for each layer of a network to reduce the number of memory accesses. DRMap~\cite{putra2020drmap} proposed a generic DRAM mapping policy and a DSE to reduce the DRAM access latency and energy. These two works were implemented on Tensor Processing Units (TPUs).~\cite{kang2021multi} defined a multi-bank on-chip memory management (MOMM) problem to minimize the DRAM
access overhead in the processing of CNNs on a neural processing unit (NPU) with a multi-bank on-chip memory. However, since FPGAs have different hardware architecture with TPUs or NPUs, their optimizing algorithms cannot be directly applied to FPGA-based designs. For example, in the TPU-based designs~\cite{putra2021romanet,putra2020drmap}, the architecture of the on-chip accelerator is already fixed, with fixed MAC arrays and fixed on-chip buffers for input features, output features, and weights, respectively. In FPGA-based designs, only the total number of DSPs and on-chip memory sizes are given, and the allocation and connection of MAC arrays and individual buffers are configured by the designer. Therefore, optimizations on FPGA-based designs should not only reduce DRAM access latency or frequency based on the off-chip DRAM access policy but also be comparable to support the on-chip acceleration designs. 

A few FPGA-based works reorganized the DRAM layout to relieve the memory access discontinuity and validated the approaches on realistic end-to-end tests. For example,~\cite{guan2017fp} compared three different layout schemes of input features in the inference phase and finally found that the channel-major scheme where the input features are fetched and stored along the input channel direction first could improve access continuity and reduce data duplication. Caffeine~\cite{zhang2018caffeine} combined both on-chip and off-chip data reorganizations for the convolutional matrix-multiplication representation to maximize the underlying memory bandwidth utilization.  FlexCNNe~\cite{sohrabizadeh2020end} further optimized data layout optimizations on the concatenation layers.  

However, all these works~\cite{guan2017fp}-\cite{sohrabizadeh2020end} are based on the computation and memory access pattern in the inference phase which only has FP. The training phase involves FP, BP, and WU where their data access pattern for output features, input features, and weights are different. Therefore, the above-mentioned approaches cannot be directly applied in CNN training, and a new optimized design considering FP, BP, and WU together is required. 

\noindent\textbf{FPGA-based Training Accelerators:} As mentioned in Section~\ref{sec:introduction}, CNN training on FPGAs has not been comprehensively investigated.} The training process is much more complicated than the inference process, so it is sub-optimal to directly adopt the frameworks of inference accelerators for training.

Due to the computation complexity and communication bottleneck, currently, only a few works aim to achieve efficient FPGA-based training. With FPGA clusters, FPDeep explored layer-level parallelism for training a CNN model in a fine-grained pipeline~\cite{wang2020fpdeep}, which has superior scalability to a large number of FPGAs. However, such larger clusters are not suitable to be adopted on edge-level applications. For training on a single FPGA, an automatic compiler for training accelerator on Stratix 10 was developed in the precision of 16-bit fixed-point~\cite{venkataramanaiah2019automatic}. DarkFPGA adopted batch-level parallelism using 8-bit integers for training a VGG-like network on the Maxeler MAX5 platform~\cite{luo2020towards}. It achieved high throughput when the batch size is large. A sparse CNN training accelerator was designed on VCU1525. The accelerator was implemented on a pre-trained CNN model with 85\% parameters pruned~\cite{nakahara2019fpga}. However, these existing works mainly focused on cloud-level devices with abundant computation and memory resources. 

Besides, even with cloud-level resources, reduced precision and pruning approaches have also been utilized to decrease computation intensity and communication bottleneck. Although quantization adopted in prior training accelerators~\cite{luo2020towards,fox2019training} led to remarkable benefits in terms of resource usage and power consumption, these works have not provided any evidence that such quantization techniques can remain high accuracy on a large dataset (e.g. ImageNet) with dense neural networks. Currently, training with full precision is still preferred in most realistic applications, and its high computation and memory overhead should be faced directly.  However, none of the above-mentioned state-of-art training accelerators targeted resource-limited edge FPGAs with full precision, which is more challenging to implement end-to-end CNN training but is more practical in real-world scenarios. Therefore, an optimized design is necessary to implement on-device training on resource-limited FPGAs without sacrificing precision.

{
\noindent\textbf{Implementation of BN layers:} Apart from the computation-intensive Conv layer, the BN layer is also a key component and is essential for the training process. In inference, a BN layer can be folded into the adjacent CONV layer, since it just performs a simple linear transformation~\cite{lu2020reconfigurable}. However, the batch normalization process in the training phase is much more complex. It needs to calculate the expected value and variance of the data in the whole mini-batch, which involves lots of on-chip and off-chip data transmission. Lu et al.~\cite{lu2020reconfigurable} optimized the computation flow of BN layers during
FP and BP, and implement BN layers in their CNN training accelerator. Unlike~\cite{lu2020reconfigurable} which adopts the 8-bit fixed-point in Conv layers and FP16 in BN layers, our work supports BN layers with full precision, which brings more challenges for computation and transmission requirements.
}

\subsection{Motivations of The Proposed Design}
\label{sec:Motivations}
To implement on-device training on resource-limited FPGAs, we need to solve the computation complexity and communication bottleneck illustrated in Fig.~\ref{fig:overview}.

\begin{figure}[h]
  \centering
  \includegraphics[width=4.0 in
  ]{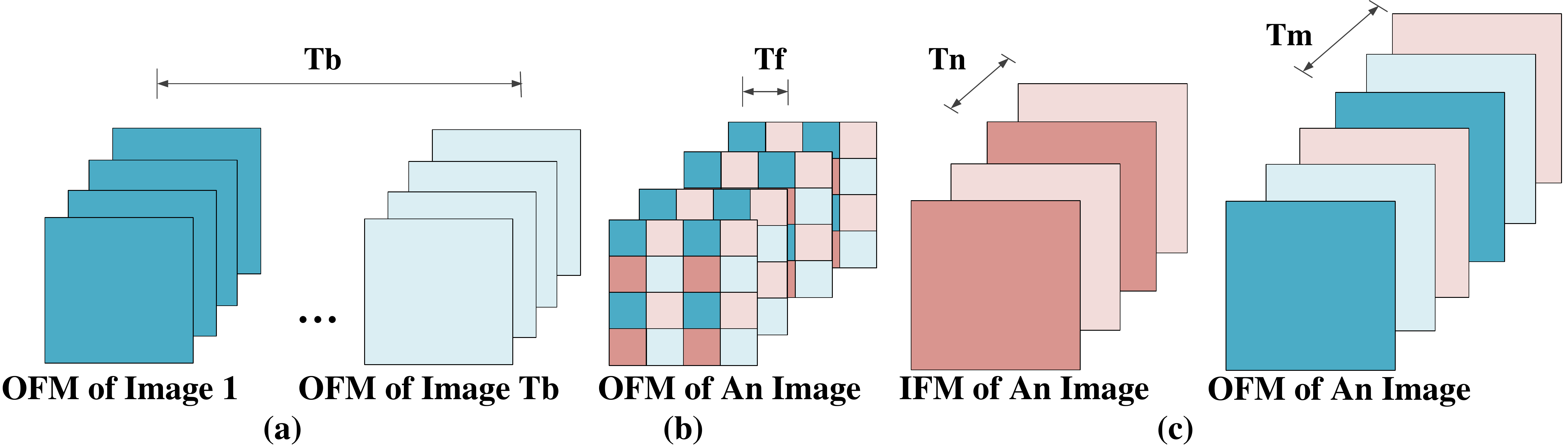}
  \caption{
  Three levels of parallelism. (a) Batch-level parallelism, (b) Feature-map-level parallelism, (c) Channel-level parallelism.}
  \label{fig:parallel}
  \Description{fig:parallel}
\end{figure}

{
In the training phase, the FP, BP, and WU processes are conducted iteratively and need to be completed on the same accelerator. Edge FPGAs have limited computational resources. Using separate kernels for FP, BP, and WU leads to resource underutilization and low energy efficiency. Therefore, to efficiently process the complex computation for FP, BP, and WU, we need to design a training accelerator that can handle the three processes in a unified convolution kernel and can achieve a high parallelism degree considering the flexibility of DNN architectures.} For a Conv layer, there are three levels of parallelism that are adopted in FPGA-based accelerators: batch-level parallelism, feature-map-level parallelism, and channel-level parallelism. {
Fig.~\ref{fig:parallel} (a) illustrates the process of batch-level parallelism, where $Tb$ nominates the number of output feature maps (OFMs) of images in a mini-batch that are processed in parallel. Fig.~\ref{fig:parallel} (b) shows the process of feature-map-level parallelism, where $Tf\times Tf$ features of OFMs are processed in parallel. Fig.~\ref{fig:parallel} (c) shows the process of channel-level parallelism, where $Tm$ nominates the number of output channels of OFMs that are processed in parallel, and $Tn$ nominates the number of input channels of input feature maps (IFMs) that are processed in parallel. The degree of parallelism depends on the amounts of utilized computation units on the hardware. Table~\ref{tab:parallel} shows the comparisons of the three levels of parallelism. Considering a Conv layer with $B$ images in a mini-batch, it is assumed that the number of input channels is $N$, the number of output channels is $M$, the size of an OFM is $R\times C$, and the size of a weights kernel is $K\times K$. $Tmops=B\times M\times N\times R\times C\times K\times K$ multiply operations are required to process such a layer. For the batch-level parallelism, it takes $\lceil\frac{
B}{Tb}\rceil\times M\times N\times R\times C\times K\times K$ cycles to complete the Conv layer. Such parallelism can achieve high throughput when the batch size is large, and the size of the feature map and the number of channels have little impact on the performance. For example, in previous works, DarkFPGA~\cite{luo2020towards} built its accelerator with batch-level parallelism and achieved high throughput when the batch size is 128. However, when the batch size is small or even 1 (online learning), most computation units will remain idle. For example, when $B<Tb$, completing the Conv layer costs $Tmops$ cycles, and $\frac{
Tb-B}{Tb}$ of computation resources remain idle. It leads to a low parallelism degree and such under-utilization of resources makes the performance sub-optimal. For the feature-map-level parallelism which has been adopted by works like~\cite{venkataramanaiah2019automatic}, it takes $B\times M\times N\times \lceil\frac{
R}{Tf}\rceil\times \lceil\frac{
C}{Tf}\rceil\times K\times K$ cycles to finish a Conv layer. The batch size and the number of channels have little impact on such parallelism. The parallelism will benefit from layers with large feature map size but has under-utilization for layers with small feature map size. For example, when $R<Tf$ and $C<Tf$, completing the Conv layer costs $Tmops$ cycles, and $\frac{Tf-R}{Tf}\times \frac{Tf-C}{Tf}$ of computation resources will remain idle. However, in CNN training, the size of a feature map may vary from large size (like $224\times 224$ for the input image of the ImageNet) to $1\times 1$ for the FC layer. The feature-map-level parallelism will be inefficient to process the layers with a small feature map. For channel-level parallelism, it takes $
B\times \lceil\frac{
M}{Tm}\rceil\times \lceil\frac{
N}{Tn}\rceil\times R\times C\times K\times K$ cycles to complete the Conv layer. It acquires a high parallelism degree with a large channel number, and the batch size and feature map size have little impact on it. When the channel number is small, for example, when $N<Tn$, completing the Conv layer costs $B\times \lceil\frac{
M}{Tm}\rceil\times N\times R\times C\times K\times K$ cycles, and $\frac{Tn-N}{Tn}$ of computation resources will remain idle. However, for most neural networks, only the first layer has a small input channel number ($N=3$). For other layers, the channel size (for example 32, 64, etc.) is usually larger than the maximum degree of parallelism that an edge FPGA can achieve. Therefore, channel-level parallelism is widely adopted by FPGA-based inference accelerators~\cite{JasonOptimizing, jiang2019achieving}. Generally speaking, the channel-level parallelism can achieve a constantly high degree of parallelism across multiple layers, so it is adopted in the proposed design as shown in Fig.~\ref{fig:overview}.} The proposed accelerator with a channel-level parallelism-based convolution kernel to process FP, BP, and WU will be introduced in detail in Section~\ref{sec:accelerator}.

\begin{table}[h]
  \caption{
  Comparisons of the three levels of parallelism}
  \label{tab:parallel}
  \begin{tabular}{cccc}
    \toprule
    Parallelism
    &\multirowcell{2}{Batch-level\\ Parallelism}
    &\multirowcell{2}{Feature-map-level\\ Parallelism}
    &\multirowcell{2}{Channel-level\\ Parallelism}
    
    \\\\
    \midrule
    Large Batch Size
    &advantaged 
    &little impact
    &little impact\\
    Small Batch Size
    &disadvantaged
    &little impact 
    &little impact
    \\  \hline
    Large Feature Map Size
    &little impact 
    &advantaged
    &little impact\\
    Small Feature Map Size
    &little impact
    &disadvantaged
    &little impact
    \\  \hline
    Large Channel Number
    &little impact 
    &little impact
    &advantaged\\
    Small Channel Number
    &little impact
    &little impact
    &disadvantaged

    \\
  \bottomrule
\end{tabular}
\end{table}

Furthermore, the communication bottleneck is also challenging for edge-level FPGAs in end-to-end training. As illustrated in Fig.~\ref{fig:training process}, the activation data in FP needs to be used in BP and WU, and the loss data generated in BP is also required in WU. The length and heterogeneity of the data dependency paths in different layers make external memory accesses inevitable~\cite{tao2020challenges}. {
Previous training accelerators attempted to avoid external memory accesses. For example, FPDeep~\cite{wang2020fpdeep} scaled CNN computations to larger clusters so that only on-chip memory is needed for the CONV layers. However, such larger clusters cannot be used on edge devices.~\cite{liu2017fpga} implemented LeNet-10 on an FPGA and stored the inputs and outputs of one layer on the chip. Such design can only support small networks, but for many larger networks (e.g. Vgg-16, AlexNet, etc.), the on-chip memory of an edge FPGA is not big enough to hold weights or features in every Conv layer. Therefore, several works~\cite{luo2020towards,fox2019training,nakahara2019fpga} applied quantization or pruning to reduce off-chip memory access. However, unlike inference where compressed networks cause little accuracy decrease~\cite{guo2017angel}, these training works have not proved that their compression techniques can remain high accuracy on large datasets with dense networks. To guarantee accuracy, it is necessary to implement CNN training with full precision. Our goal is to design a general accelerator supporting end-to-end training with both dense and small networks without sacrificing precision, so it is necessary to appropriately manage external memory access and allocate on-chip buffers. As mentioned in Section~\ref{sec:Related works}, the tiling schemes involved in on-chip accelerator design break the continuity of data addresses in DRAM and thus reduce the DMA transmission efficiency.} Therefore, it is necessary to improve the address continuity of data to improve the efficiency of data swapping considering the complex data patterns in FP, BP, and WU altogether. To solve this communication issue, as shown in Fig.~\ref{fig:overview}, a data reshaping approach is proposed and will be introduced in detail in Section~\ref{sec:memory}.

\section{FPGA-based CNN Training Accelerator}
\label{sec:accelerator}
In this section, we propose an FPGA-based accelerator exploiting channel-level parallelism to deal with the training process. A unified convolution kernel is designed to process FP, BP, and WU with full precision. 

\subsection{The Architecture of The Training Accelerator}
\label{sec:architecture}

\begin{figure}[h]
  \centering
  \includegraphics[width=5.2in
  ]
  {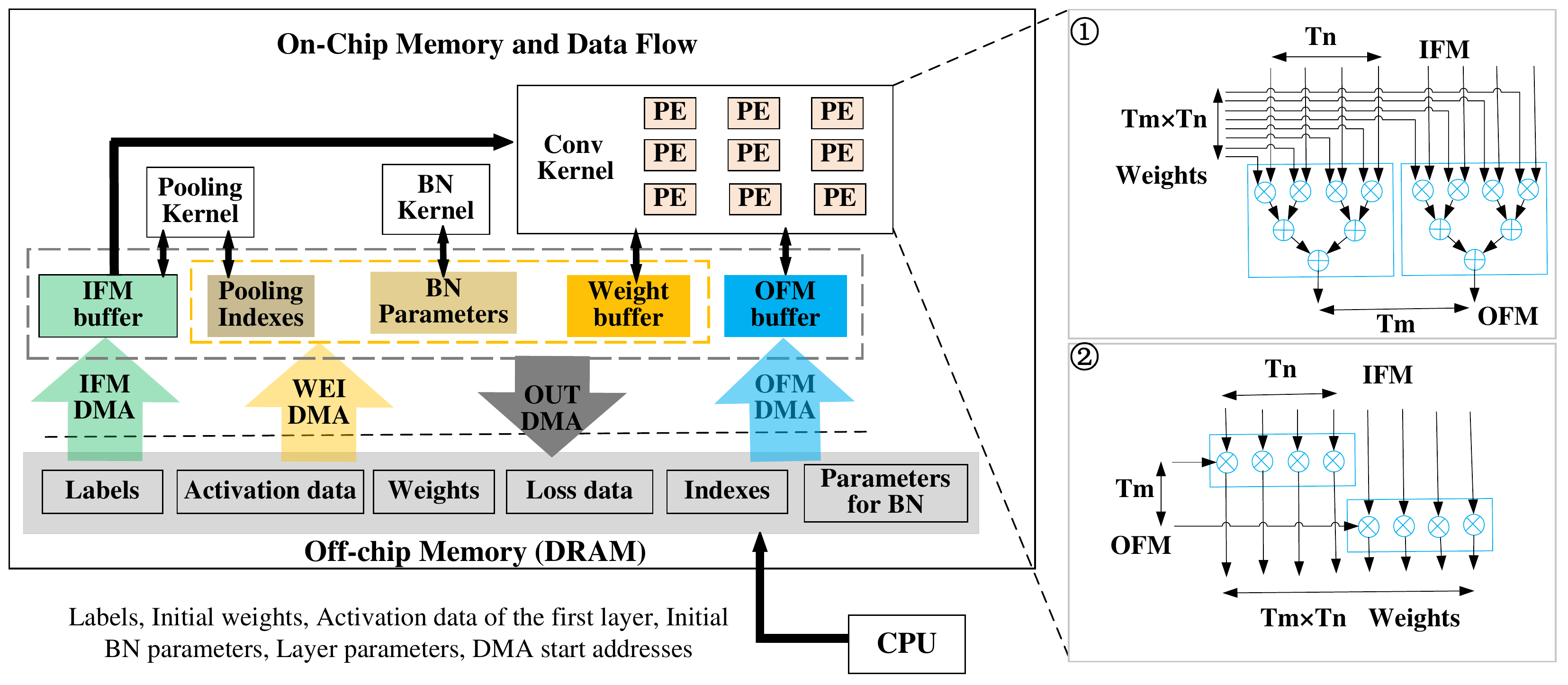}
  \caption{
  Accelerator architecture.}
  \label{fig:accelerator}
  \Description{fig:accelerator}
\end{figure}

The proposed training accelerator is shown in Fig.~\ref{fig:accelerator}. We implement our accelerator on an end-to-end training system. {
At first, the CPU transmits labels, initial weights, the activation data of the first layer, layer parameters, initial parameters for BN layers, and the DMA start addresses for each layer to the off-chip DRAM. The layer parameters include computation type (e.g. Conv, ReLU, BN, or pooling) and the shape information.} The DMA start addresses are calculated off-line according to the off-chip memory layout based on our data reshaping approach mentioned in Section~\ref{sec:memory}. Our accelerator executes computation-intensive kernels based on data dependencies within a CNN model, while the entropy loss function is computed on the off-chip ARM core. 

{
As illustrated in Fig.~\ref{fig:accelerator}, the off-chip memory (DRAM) stores data for activation, loss, weights, labels, indexes for pooling, and parameters for BN.} {
Data are transmitted through the DMA AXI-stream bus to on-chip memory for computation.  There are 4 DMA stream channels: IFM DMA, OFM DMA, WEI DMA, and OUT DMA. These 4 channels are independent and can work in parallel.} On the FPGA chip, a unified Conv kernel is designed to process FP, BP, and WU with the same computation resources, i.e. the digital signal processors (DSPs). {
The Conv kernel is composed of multiple processing elements (PEs) to implement MAC operations. The adder tree structure is adopted for the proposed kernel since it is flexible to support different computation patterns for FP, BP, and WU. The connection of each multiplier and adder for FP and BP is shown in \textcircled{1} of Fig.~\ref{fig:accelerator}, while the connection of each multiplier and adder for WU is shown in \textcircled{2} of Fig.~\ref{fig:accelerator}.} The Pooling Kernel focuses on the pooling operation. {
The BN Kernel achieves batch normalization and updates BN parameters during FP and BP. ReLU is always followed by a Conv or BN layer. The accelerator compares the output features with 0 when storing output features back to the DRAM in Conv or BN layers, so ReLU does not need a unique functional unit. Five types of on-chip block RAMs (BRAMs) are used to buffer IFMs, weights or weights gradients, OFMs, pooling indexes, and BN parameters.} We adopt double-buffer designs so that data transmission and computation can be conducted in parallel. 

\subsection{The Forward and Backward Propagation of A Convolutional Layer}
\label{FP/BP}

\begin{table*}
  \caption{Definitions of Symbols}
  \label{tab:symbols}
  \begin{tabular}{cl}
    \toprule
    Notation 
    &Description\\
    \midrule
    $i$,$j$ & Index of a Conv layer\\
    $B$ & Batch size\\
    $N^i$ & Number of the input channels of the $i$th Conv layer\\
    $M^i$ & Number of the output channels of the $i$th Conv layer\\
    $R^i$ & Number of the rows of the OFMs for the $i$th Conv layer\\
    $C^i$ & Number of the columns of the OFMs for the $i$th Conv layer\\
    $K^i$ & Size of the weights kernel for the $i$th Conv layer\\
    $S^i$ & Stride for the $i$th Conv layer\\
    $A_{i}[b,m,r,c]$ & Activation for the $i$th Conv layer in FP\\
    $L_{i}[b,n,r,c]$ & Loss for the $i$th Conv layer in BP\\
    $W_{i}[n,m,kr,kc]$ & Weights for the $i$th Conv layer\\
    $dW_{i}[n,m,kr,kc]$ & Weights gradients for the $i$th Conv layer in WU\\
    $Tm$ & Number of the output channels in a tile of output features\\& in each Conv layer\\
    $Tn$ & Number of the input channels in a tile of input features in \\&each Conv layer\\
    $Tr^i$ & Number of the rows in a tile of output features in the $i$th Conv layer\\
    $Tc^i$ & Number of the columns in a tile of output features in the $i$th
    Conv layer\\
    $M^i\_{on}$ &Number of the output channels of the weights that stored on-chip in 
    \\&the $i$th Conv layer\\
    
    $R^j\_in$ & Number of the rows of the IFMs for the $j$th Conv layer\\
    $C^j\_in$ & Number of the columns of the IFMs for the $j$th Conv layer\\
    $Tr^j\_in$ & Number of the rows in a tile of input features in the $j$th Conv layer\\
    $Tc^j\_in$ & Number of the columns in a tile of input features in the $j$th Conv layer\\
    
  \bottomrule
\end{tabular}
\end{table*}

Our accelerator adopts channel-level parallelism, loop unrolling, and loop tiling. The symbols are defined in Table~\ref{tab:symbols}. In channel-level parallelism, $Tn$ and $Tm$ are determined by available computation resources (i.e. DSPs) on the FPGA chip and are fixed for all Conv layers. The degree of parallelism is determined by $Tn\times Tm$.

Our accelerator achieves SGD in CNN training. The forward and backward propagation processes of a Conv layer processing the $b$th image in a mini-batch can be formulated in Eq.~\eqref{eq:FPconv} and~\eqref{eq:BPconv}, where $W'_i$ is the transposed and flipped tensor of $W_i$. {
As illustrated in \textcircled{1} of Fig.~\ref{fig:accelerator}, in FP and BP, the Conv Kernel conducts MAC operations for weights and input features from activation or loss.} The IFM buffer stores a tile of activation or loss transmitted via the IFM DMA channel, and the Weight buffer stores weights transmitted via the WEI DMA channel. The OFM buffer stores a tiled of MAC outcomes. Computation results are transmitted to the DRAM via the OUT DMA channel. If a Conv layer is followed by a ReLU layer, for FP, the data in the OFM buffer will be compared with 0 before entering into the OUT DMA channel. For BP, the activation of the previous layer will be transmitted via the OFM DMA channel, and the accelerator decides which value should be propagated according to Eq.~\eqref{eq:BPrelu}. 
\begin{equation}
    A_{i+1}[b,m,r,c] =\sum_{n=1}^{N^i}\sum_{kr=1}^{K^i}\sum_{kc=1}^{K^i} A_{i}[b,n,S^i\times r+kr,S^i\times c+kc]\times W_{i}[m,n,kr,kc] 
\label{eq:FPconv}
\end{equation}

\begin{equation}
    L_{i}[b,n,r,c] =\sum_{m=1}^{M^i}\sum_{kr'=1}^{K^i}\sum_{kc'=1}^{K^i} L_{i+1}[b,m,S^i\times r+kr',S^i\times c+kc']\times W'_{i}[n,m,kr',kc']
\label{eq:BPconv}
\end{equation}

\begin{equation}
    L_{i}[b,m,r,c]=
    \begin{cases} L_{i+1}[b,m,r,c], & A_{i}[b,m,r,c]>0,\\
    0, & others
    \end{cases}
\label{eq:BPrelu}
\end{equation}

\subsection{The Weight Update of A Convolutional Layer}
\label{WU}

The gradients of weights can be calculated in Eq.~\eqref{eq:WUconv}. {
The generated hardware implementation of the PE architecture processing WU operations in the Conv kernel is shown in \textcircled{2} of Fig.~\ref{fig:accelerator}.} During the WU, the Conv Kernel conducts MAC operations for the activation data transmitted via the IFM DMA channel and the loss data transmitted via the OFM DMA channel. The gradients are stored in the Weight buffer. Once the Conv Kernel completes the computation for the last image in a mini-batch, the original weights are transmitted via the WEI DMA channel. Then, weights are updated by deducting the product of the gradients and learning rate. New weights are sent back to DRAM via the OUT DMA channel.  
\begin{equation}
    dW_{i}[m,n,kr,kc] =\sum_{b=1}^{B}\sum_{r=1}^{R^i}\sum_{c=1}^{C^i} L_{i+1}[b,m,r,c]\times A_{i}[b,n,S^i\times r+kr,S^i\times c+kc]
\label{eq:WUconv}
\end{equation}

\subsection{The Forward and Backward Propagation of A Pooling Layer}
\label{Pool}

In the FP process of a pooling layer, the activation is transmitted via the IFM DMA channel and stored in the IFM buffer. In the maximum pooling, the Pooling Kernel compares adjacent pixels, transfers the results back to DRAM via the OUT DMA channel, and records the index for the maximal pixel into the Pooling Indexes buffer. The index of a pixel is a 2-bit integer. For average pooling, the kernel just calculates the average value of a patch of features. In the BP process of maximum pooling, the indexes are loaded back via the WEI DMA channel, and loss from the previous layer is loaded via the IFM DMA channel. The Pooling Kernel compares the indexes and stores the propagated value into the IFM buffer. The BP process of the maximum pooling is formulated in Eq.~\eqref{eq:BPpool}. For average pooling, the loss values of a patch are directly accumulated. After a tile of data is processed, the calculated loss is sent back via the OUT DMA channel.

\begin{equation}
    L_{i}[b,m,S^i\times r+kr,S^i\times c+kc]=
\begin{cases}
L_{i+1}[b,m,r,c], A_{i+1}[b,m,r,c]= A_{i}[b,m,S^i\times r+kr,S^i\times c+kc],\\ 0, others
\end{cases}
\label{eq:BPpool}
\end{equation}

{
\subsection{The Forward Propagation of A BN Layer}
\label{BNFP}

Our BN kernel is based on the computation flow in~\cite{lu2020reconfigurable}. However, unlike the prior work which utilizes half-precision, we adopt full precision which brings more computation and transmission challenges. The BN parameters in Fig.~\ref{fig:accelerator} includes learnable parameters $\gamma_{i}[m]$ and $\beta_{i}[m]$ and immediate parameters $\lambda_i[m]$ and $\hat A_i[b,m,r,c]$, where $m$ is the index of the channel. The $\gamma_{i}[m]$ and $\beta_{i}[m]$ are used to generate the immediate parameters and the output activation $A_{i+1}[b,m,r,c]$ during FP. During BP, the immediate parameters and the loss propagated from the next layer $L_{i+1}[b,m,r,c]$ are used to update the learn parameters and propagate the loss $L_{i}[b,m,r,c]$ back. Since the size of $\gamma_{i}[m]$, $\beta_{i}[m]$, and $\lambda_i[m]$ is $M$ (the number of the output channels), the on-chip BRAMs are large enough to hold these data in a BN layer. Therefore, we use the BN Parameters buffer to store these parameters as well as the expected value and variance. The $\hat A_i[b,m,r,c]$ is transmitted to DRAM together with $A_{i+1}[b,m,r,c]$.

In FP, the BN Kernel first loads $\gamma_{i}[m]$ and $\beta_{i}[m]$ from DRAM to the BN Parameters buffer via the WEI DMA channel. Then it loads $A_{i}[b,m,r,c]$ via the IFM DMA channel and calculates the expected value $E(X)_{i}[m]$ and variance $V(X)_{i}[m]$ according to Eq.~\eqref{eq:BNEX}-\eqref{eq:BNVX}. To avoid disarranging the DRAM data layout for adjacent Conv layers, we load data tile by tile using the same data format as that in Conv layers. The expected value and variance are calculated after the entire data of a mini-batch is accessed. Then input activation is loaded from the beginning to calculate the immediate parameters according to Eq.~\eqref{eq:BNlambda} and Eq.~\eqref{eq:BNhat}, where $\epsilon$ is a constant parameter. $\lambda_{i}[m]$ is stored in the BN Parameters buffer, while $\hat A_i[b,m,r,c]$ is transmitted to DRAM via the OUT channel in parallel with activation loading. Finally, the output activation is calculated according to Eq.~\eqref{eq:BNFP}. The BN operation completes after the activation, $\gamma_{i}[m]$, $\beta_{i}[m]$, and $\lambda_i[m]$ are stored to DRAM.  
\begin{equation}
    E(X)_{i}[m]=\frac{1}{B\times R^i \times C^i}\sum_{b=1}^{B}\sum_{r=1}^{R^i}\sum_{c=1}^{C^i}
 A_{i}[b,m,r,c]
\label{eq:BNEX}
\end{equation}

\begin{equation}
    E(X^2)_{i}[m]=\frac{1}{B\times R^i \times C^i}\sum_{b=1}^{B}\sum_{r=1}^{R^i}\sum_{c=1}^{C^i}
 A^2_{i}[b,m,r,c]
\label{eq:BNEX2}
\end{equation}

\begin{equation}
     V(X)_{i}[m]=E(X^2)_{i}[m]-(E(X)_{i}[m])^2
\label{eq:BNVX}
\end{equation}

\begin{equation}
     \lambda_{i}[m]=\frac{1}{\sqrt{V(X)_{i}[m]+\epsilon}}
\label{eq:BNlambda}
\end{equation}

\begin{equation}
     \hat A_i[b,m,r,c]= (A_i[b,m,r,c]-E(X)_{i}[m])\times\lambda_{i}[m] 
\label{eq:BNhat}
\end{equation}

\begin{equation}
     A_{i+1}[b,m,r,c]= \hat A_i[b,m,r,c]\times\gamma_{i}[m]+\beta_{i}[m] 
\label{eq:BNFP}
\end{equation}

\subsection{The Backward Propagation of A BN Layer}
\label{BNBP}

In BP, $\hat A_i[b,m,r,c]$, $\lambda_{i}[m]$ and $L_{i+1}[b,m,r,c]$ are used to update the learnable parameters $\gamma_{i}[m]$ and $\beta_{i}[m]$, and $L_{i}[b,m,r,c]$ is propagated back. $\lambda_{i}[m]$, $\gamma_{i}[m]$, and $\beta_{i}[m]$ are first uploaded via the WEI channel and stored in the BN Parameters buffer. Then, the BN Kernel loads $\hat A_i[b,m,r,c]$ and $L_{i+1}[b,m,r,c]$ via the IFM and OFM channel respectively to calculate the gradients for $\gamma_{i}[m]$ and $\beta_{i}[m]$ according to Eq.~\eqref{eq:BNdgamma} and~\eqref{eq:BNdbeta}. The learnable parameters are updated by deducting the gradients, while $L_i[b,m,r,c]$ is calculated according to Eq.~\eqref{eq:BNBP}. 
\begin{equation}
     d\gamma_{i}[m]=\sum_{b=1}^{B}\sum_{r=1}^{R^i}\sum_{c=1}^{C^i}
 L_{i+1}[b,m,r,c]\times\hat A_i[b,m,r,c]
\label{eq:BNdgamma}
\end{equation}

\begin{equation}
     d\beta_{i}[m]=\sum_{b=1}^{B}\sum_{r=1}^{R^i}\sum_{c=1}^{C^i}
 L_{i+1}[b,m,r,c]
\label{eq:BNdbeta}
\end{equation}

\begin{equation}
     L_i[b,m,r,c]=\gamma_{i}[m]\times\lambda_{i}[m]\times(L_{i+1}[b,m,r,c]-\frac{d\beta_{i}[m]}{B\times R^i \times C^i}-\hat A_i[b,m,r,c]\times\frac{d\gamma_{i}[m]}{B\times R^i \times C^i})
\label{eq:BNBP}
\end{equation}
}

\section{Data Reshaping Approach}
\label{sec:memory}

In this section, we propose 
a data reshaping approach to solve the communication bottleneck between the on-chip buffer and off-chip memory in realistic end-to-end training processes. {
We first analyze the discontinuous memory access for the isolate accelerator with the unified channel-level parallelism-based Conv kernel proposed in Section~\ref{sec:accelerator}. Then we introduce our data reshaping approach which involves three aspects. We first achieve intra-tile continuous memory allocation by reorganizing the DRAM layouts for input features, output features, and weights. Then, we re-schedule the loop order to achieve inter-tile continuous memory allocation. These two parts are optimized together. Finally, considering the training process involves convolution operations among a mini-batch, we propose and apply a weight reuse strategy based on the proposed data layout.}

\subsection{Analysis on Discontinuous Memory Access}
\label{sec:Memory Analysis}
\begin{figure}[h]
  \centering
  \includegraphics[width=4.0 in
  ]{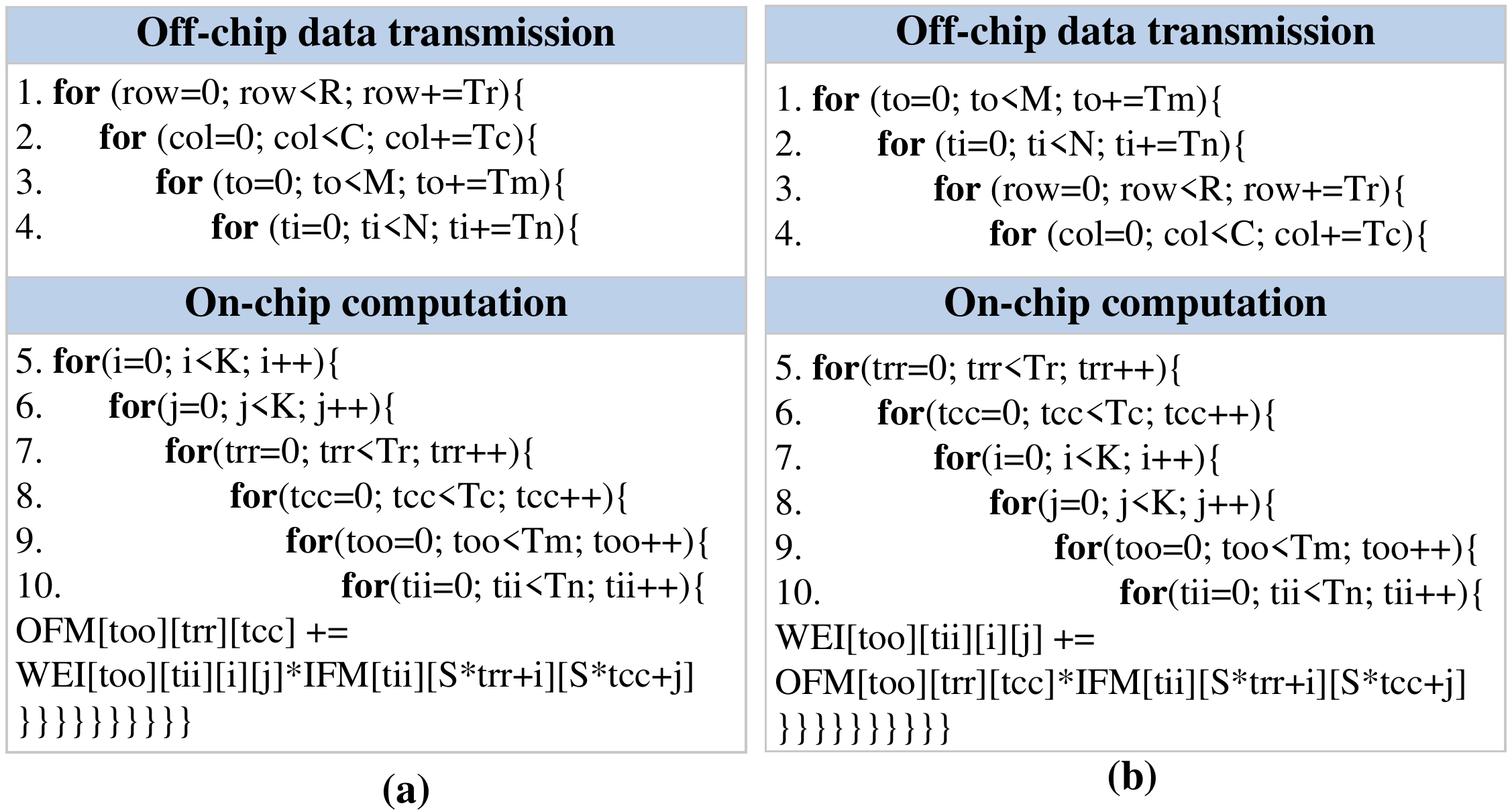}
  \caption{
  Pseudo-code of a tiled convolution layer. (a) Pseudo-code for FP and BP, (b) Pseudo-code for WU.}
  \label{fig:looporig}
  \Description{fig:looporig}
\end{figure}

The pseudo-code of a tiled convolution layer is shown in Fig.~\ref{fig:looporig}. The pseudo-code in Fig.~\ref{fig:looporig} (a) is applied in FP and BP, following previous FPGA-based inference works~\cite{JasonOptimizing,zhang2018caffeine}, while the pseudo-code in Fig.~\ref{fig:looporig} (b) is applied in WU, based on the accelerator design proposed in Section~\ref{sec:accelerator}. As discussed in Section~\ref{sec:Background}, the continuity of data significantly influences the DMA transmission efficiency. In this section, we analyze the data discontinuity when features are placed with the BCHW pattern and the BHWC pattern, where B represents batch, C represents channel, H represents height (row), and W represents width (column).

\begin{figure}[h]
  \centering
  \includegraphics[width=4.6 in
  ]{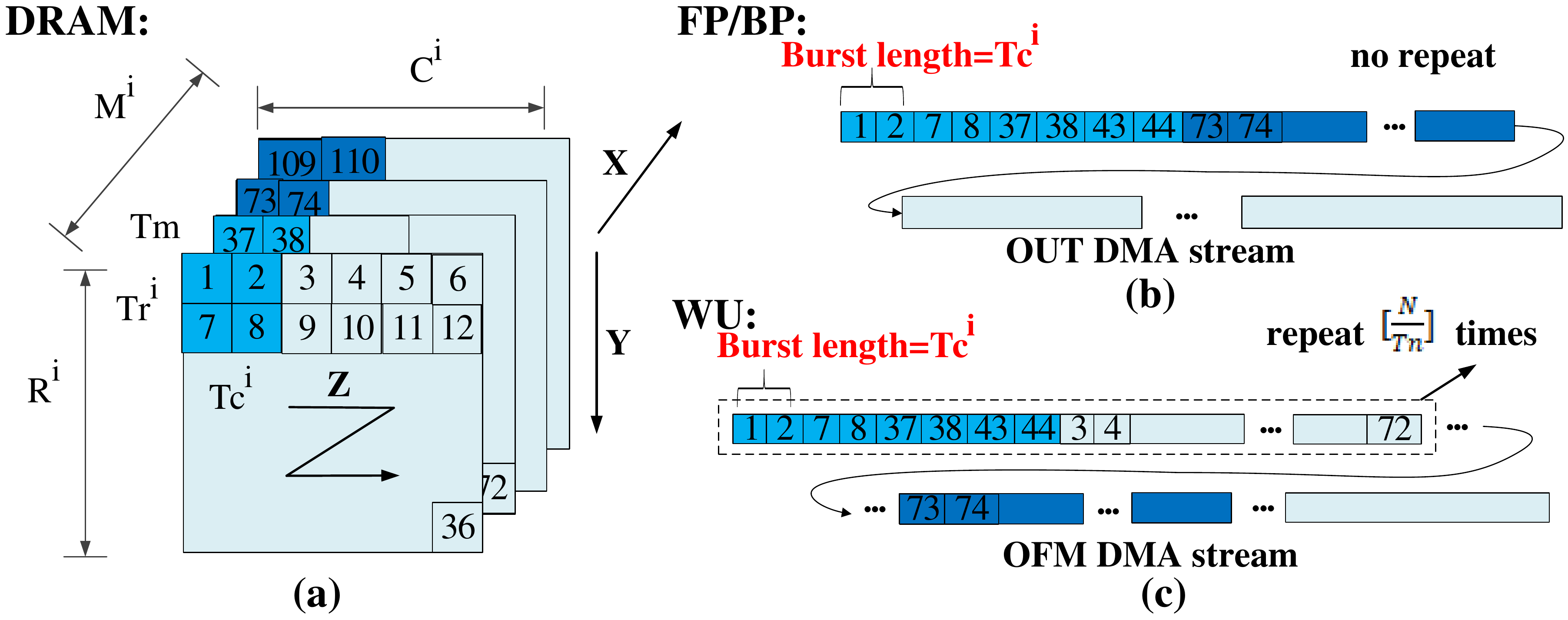}
  \caption{
  Data layout of output features before reshaping. (a) Data stored in DRAM, (b) Data transmitted via the OUT DMA channel in FP/BP, (c) Data transmitted via the OFM DMA channel in WU.}
  \label{fig:OFMbefore}
  \Description{fig:OFMbefore}

  \includegraphics[width=5.4 in
  ]{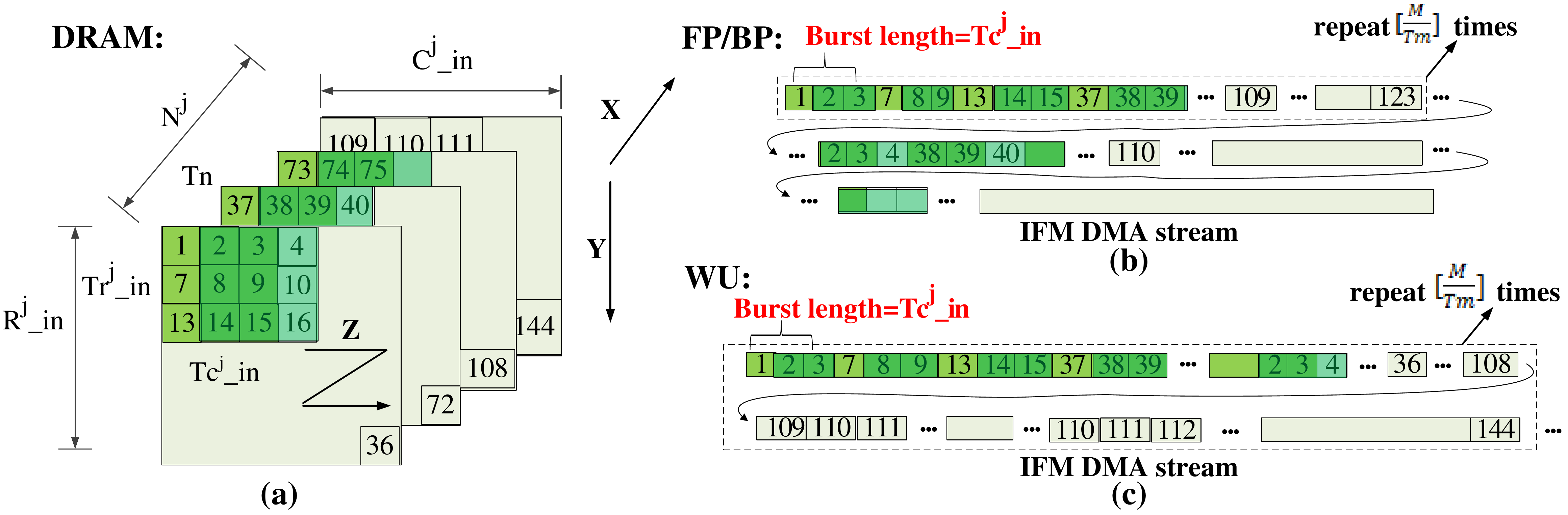}
  \caption{
  Data layout of input features before reshaping. (a) Data stored in DRAM, (b) Data transmitted via the IFM DMA channel in FP/BP, (c) Data transmitted via the IFM DMA channel in WU.}
  \label{fig:IFMbefore}
  \Description{fig:IFMbefore}

  \includegraphics[width=4.8 in
  ]{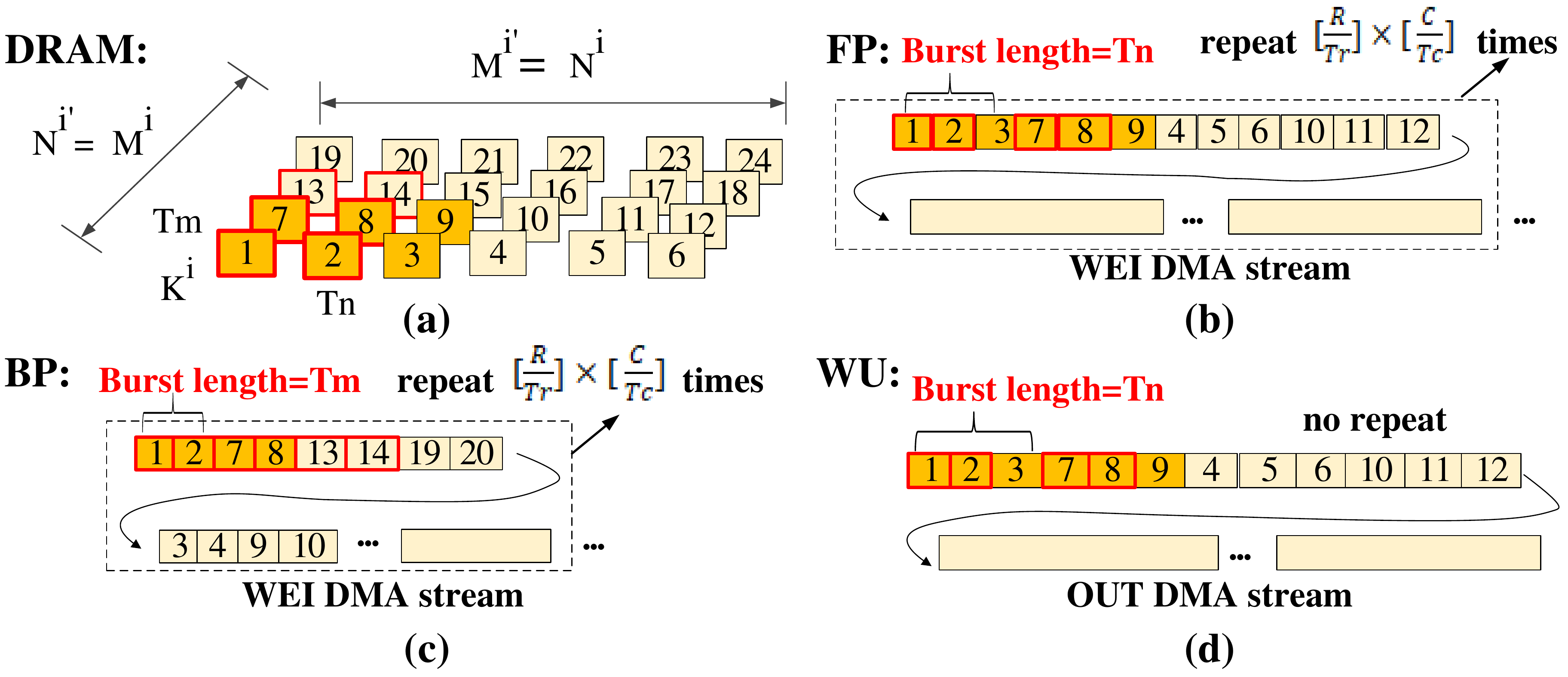}
  \caption{
  Data layout of weights before reshaping. (a) Weights stored in DRAM, (b) Weights transmitted via the WEI DMA channel in FP, (c) Weights transmitted via the WEI DMA channel in BP, (d) Weights transmitted via the OUT DMA channel in WU.}
  \label{fig:WEIbefore}
  \Description{fig:WEIbefore}
\end{figure}

\noindent\textbf{Features are placed in the BCHW pattern:}  Fig.~\ref{fig:OFMbefore} (a) shows the data layout of $M^i\times R^i\times C^i$ output features stored in DRAM for the $i$th layer. {
The output features are placed with the BCHW pattern commonly used in CNN accelerating CPU, and GPU platforms~\cite{chetlur2014cudnn,OpenVINO}. While OpenVINO~\cite{OpenVINO} is primarily for CPUs, it would also work for CPU, GPU, and FPGA platforms.} In this layout, a cube represents an element of the features, and the indexes represent the orders of the elements stored in DRAM. In FPGA-based DNN deployments, data are fetched and processed in tiles. As shown in Fig.~\ref{fig:OFMbefore}, the size of a tile is $Tm\times\!Tr^i \times Tc^i$ for output features.

The output features of layer $i$ are also the input features for its next layer $j$. As shown in Fig.~\ref{fig:IFMbefore} (a), the size of the input features in layer $j$ is $N^j\times R^j\_in\times C^j\_in$. For input features, the size of a tile is $Tn\times Tr^j\_in \times Tc^j\_in$.

{
In each DMA stream, the AXI-stream bus allows a pipeline data stream when the data addresses are continuous. Burst length represents the number of data with continuous addresses in the data stream. When a discontinuity happens, the DMA needs to be restarted. Therefore, our goal is to avoid discontinuity, i.e. elongate the burst length for different transmission patterns.}

During FP, the Conv Kernel conducts MAC operations with weights and input features, and then it generates output features. The output features are transmitted to the DRAM via the OUT DMA channel, which is shown in Fig.~\ref{fig:OFMbefore} (b). For the next layer, the input features are fetched from DRAM to the FPGA chip via the IFM DMA stream, which is shown in Fig.~\ref{fig:IFMbefore} (b). {
As illustrated in Fig.~\ref{fig:looporig} (a), in a Conv layer, the OFM buffer is reused to store and accumulate the immediate convolution results between each tile of input features and each tile of weights.} 
The first tile of output features is generated by accumulating the convolution results when the input features tiles move from the first input channel to the last channel. It corresponds to the movement in the~\textbf{X} direction in Fig.~\ref{fig:IFMbefore}. Then, the next tiles of output features are generated in the~\textbf{X} direction in Fig.~\ref{fig:OFMbefore}, so the data access pattern of input features (the dashed box in Fig.~\ref{fig:IFMbefore} (b)) repeats $\lceil\frac{
M}{Tm}\rceil$ times. After the output features tiles move from the first output channel to the last output channel, they begin to move in the~\textbf{Z} direction, and the input features tiles follow the~\textbf{Z} direction as well. 
From Fig.~\ref{fig:OFMbefore} (b) and Fig.~\ref{fig:IFMbefore} (b), the address of data is discontinuous for both inside and outside of a tile. The burst length of the output features in the $i$th layer is $Tc^i$, and the burst length of the input features in the $j$th layer is $Tc^j\_in$. The data movement in BP is similar to that in FP.

{
In WU, the data access pattern inside a tile is the same as that in FP/BP. However, the inter-tile data access pattern is different from that for FP/BP. It is because, in WU, the Conv Kernel conducts MAC operations for input features (the activation data transmitted via the IFM DMA channel) and output features (the loss data transmitted via the OFM DMA channel) to calculate weight gradients. {
Therefore, as shown in Fig.~\ref{fig:looporig} (b), the WEI buffer is reused to store and accumulate the immediate convolution result between each tile of input features and each tile of output features.} 
The first tile of weight gradients is generated when the input and output features tiles move from the first row and the first column to the last row and the last column. It corresponds to the movement in the~\textbf{Z} direction in Fig.~\ref{fig:OFMbefore} and Fig.~\ref{fig:IFMbefore}. Then, the next tile of weight gradients is generated along the input channel direction. The tiles of input features move along the~\textbf{X} direction, while the pattern of output feature tiles (the dashed box in Fig.~\ref{fig:OFMbefore} (c)) repeats $\lceil\frac{
N}{Tn}\rceil$ times. After the gradients of weights are calculated from the first input channel to the last input channel, the next tiles are generated along the output channel direction. Thus, the pattern of input feature tiles (the dashed box in Fig.~\ref{fig:IFMbefore} (c)) repeats  $\lceil\frac{
M}{Tm}\rceil$ times, while the output feature tiles move along the~\textbf{X} direction. As shown in Fig.~\ref{fig:IFMbefore} (c), the burst length for input features is $Tc^j\_in$. As shown in Fig.~\ref{fig:OFMbefore} (c), the burst length for output features is $Tc^i$.}

In CNN training, the data layout of weights is also more complex compared to the inference process. As illustrated in Fig.~\ref{fig:WEIbefore} (a), $M^i\times N^i\times K^i\times K^i$ weights of layer $i$ are stored in DRAM. {
In FP, weights are fetched in the input channel first and then the output channel when the output features are generated along the~\textbf{X} direction in Fig.~\ref{fig:OFMbefore}. Then, the output features are generated along the~\textbf{Z} direction, while the weights access pattern (the dashed box in Fig.~\ref{fig:WEIbefore} (a)) repeats $\lceil\frac{
R}{Tr}\rceil\times\lceil\frac{
C}{Tc}\rceil$ times. WU shares the same intra-tile weights access pattern with FP, but it does not need to repeat during inter-tile data access. The burst lengths for FP and WU are both $Tn$. In BP, each $K^i\times K^i$ kernel needs to be flipped. Such reallocation can be processed on the FPGA chip. However, since the numbers of input channels and output channels are interchanged, the memory access pattern of a tile is also changed. The weights kernels are transposed between the input channel dimension and the output channel dimension.} In Fig.~\ref{fig:WEIbefore}, the yellow cubes represent a tile of weights in FP and WU, and the cubes with the red box represent a tile of weights in BP. As illustrated in Fig.~\ref{fig:WEIbefore} (c), in BP, the number of output channels becomes $M^{i'}=N^i$, the number of input channels becomes $N^{i'}=M^i$, and the burst length is $Tm$.

\begin{figure}[h]
  \centering
  \includegraphics[width=4.4 in
  ]{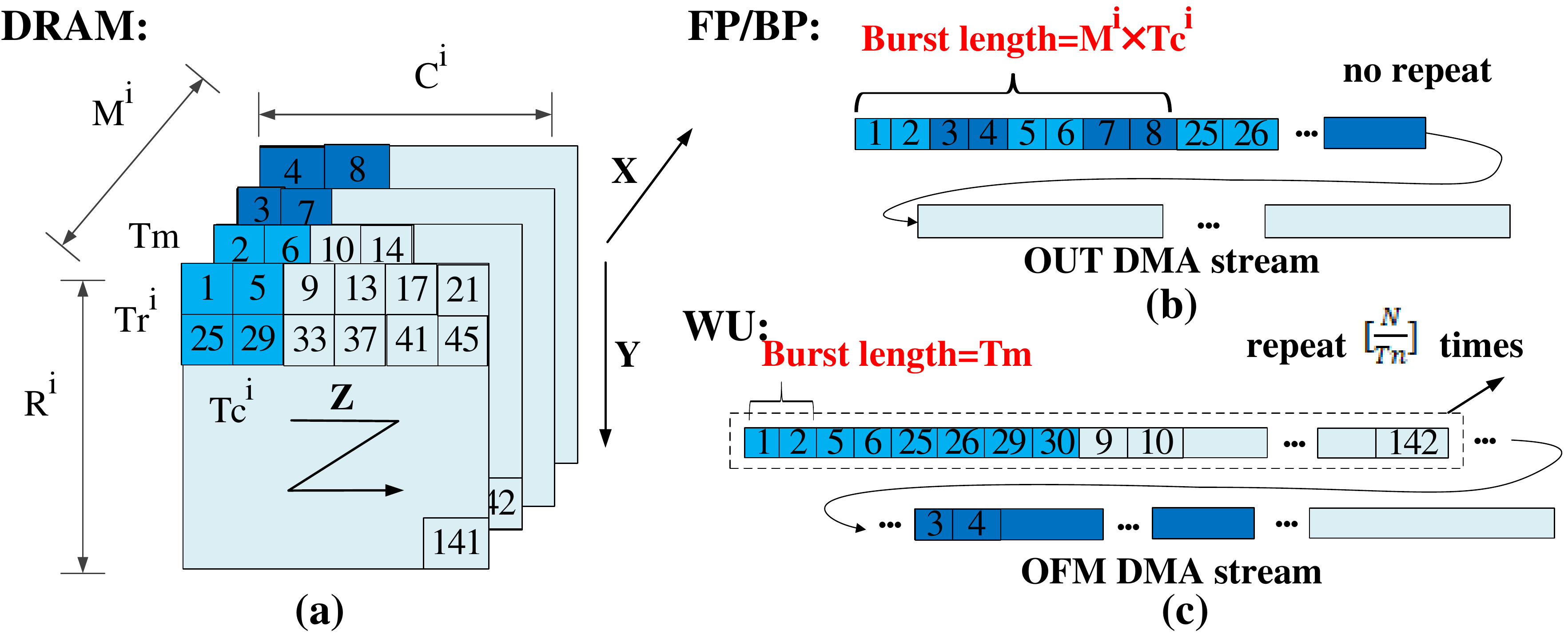}
  \caption{
  Data layout of output features with the BHWC memory allocation and feature reuse. (a) Data stored in DRAM, (b) Data transmitted via the OUT DMA channel in FP/BP, (c) Data transmitted via the OFM DMA channel in WU.}
  \label{fig:OFMchannel}
  \Description{fig:OFMchannel}

  \includegraphics[width=5.4 in
  ]{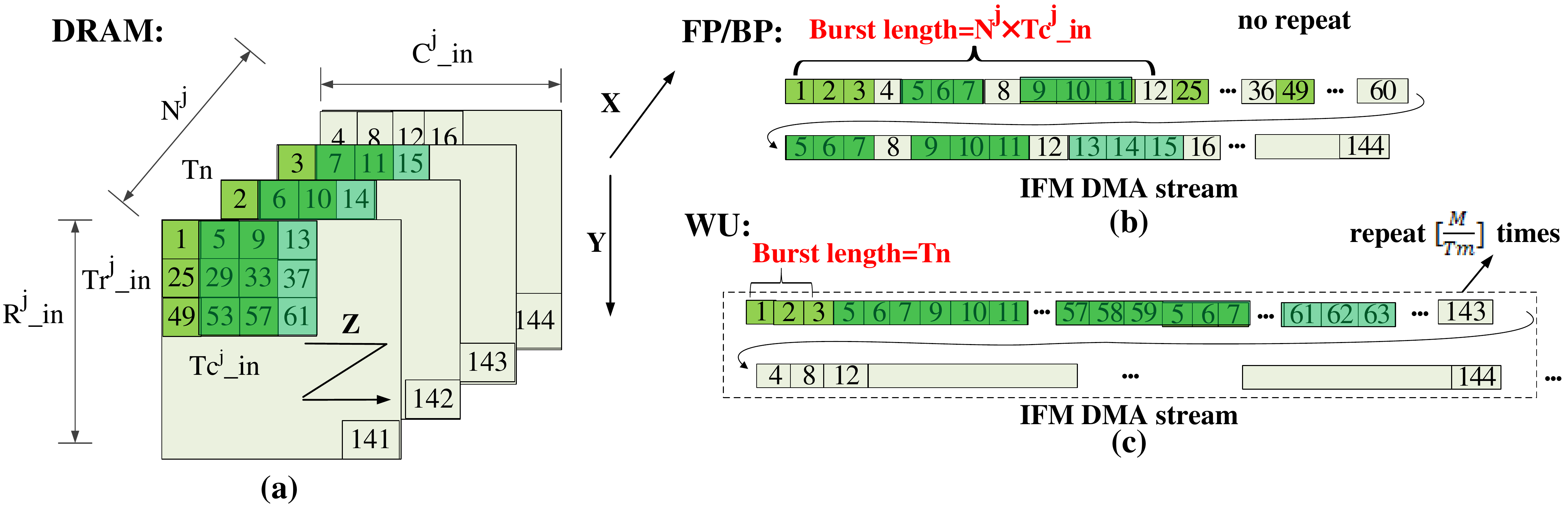}
  \caption{
  Data layout of input features with the BHWC memory allocation and feature reuse. (a) Data stored in DRAM, (b) Data transmitted via the IFM DMA channel in FP/BP, (c) Data transmitted via the IFM DMA channel in WU.}
  \label{fig:IFMchannel}
  \Description{fig:IFMchannel}

  \includegraphics[width=4.8 in
  ]{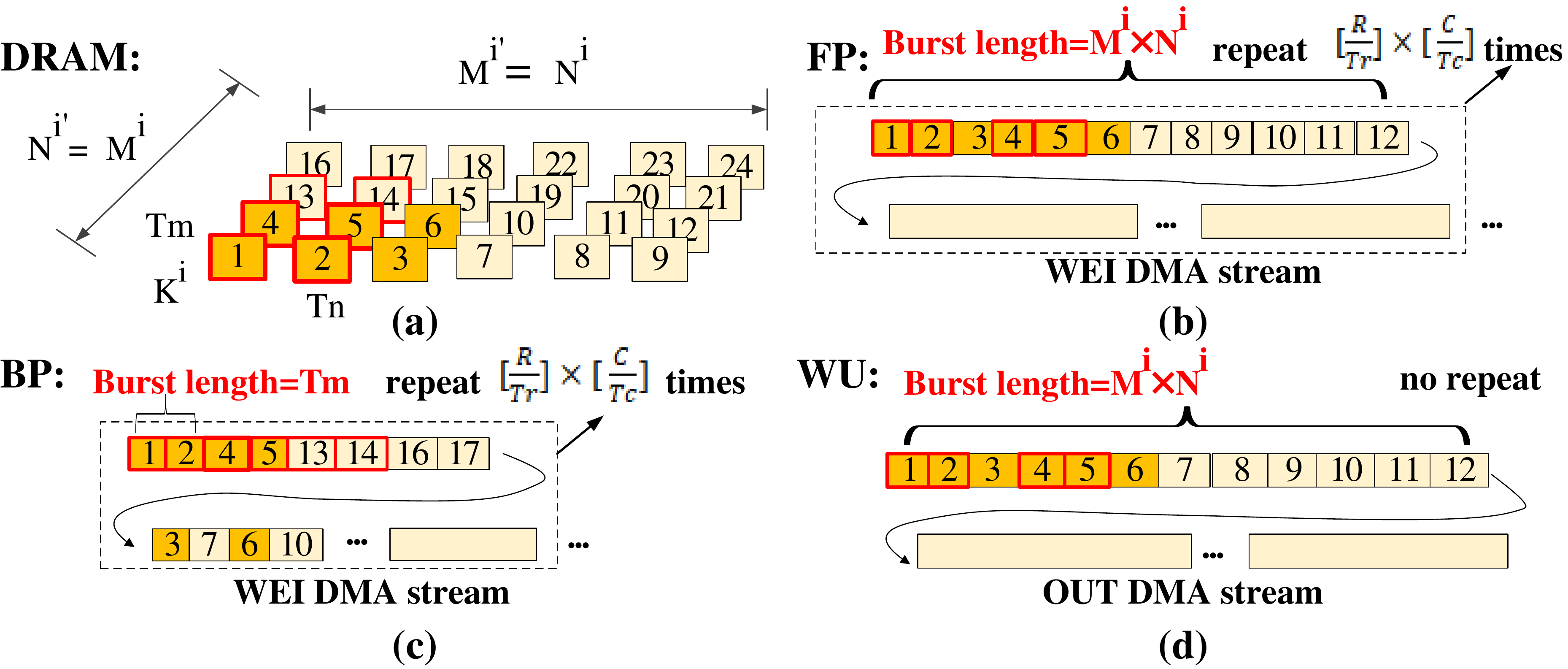}
  \caption{
  Data layout of weights placed tile by tile based on inference-based data flow. (a) Weights stored in DRAM, (b) Weights transmitted via the WEI DMA channel in FP, (c) Weights transmitted via the WEI DMA channel in BP, (d) Weights transmitted via the OUT DMA channel in WU.}
  \label{fig:WEIchannel}
  \Description{fig:WEIchannel}
\end{figure}

\noindent\textbf{Features are placed in the BHWC pattern:}  
As can be seen from Fig.~\ref{fig:OFMbefore}, Fig.~\ref{fig:IFMbefore}, and Fig.~\ref{fig:WEIbefore}, the tiled data breaks the data continuity of memory access in FP, BP, and WU for the isolate accelerator. In FPGA-based inference works, the BHWC pattern is also commonly used in end-to-end designs to optimize memory access~\cite{guan2017fp,zhang2018caffeine}. Fig.~\ref{fig:OFMchannel} (a) and Fig.~\ref{fig:IFMchannel} (a) show the data layout of features placed in the BHWC pattern following previous inference-based works. According to the loop order in Fig.~\ref{fig:looporig} (a), in FP and BP, tiles move in the channel dimension first and then move in the~\textbf{Z} direction. Therefore, it is effective to fetch $\lceil\frac{N}{Tn}\rceil$ tiles of input features to the on-chip memory and reuse the data after $\lceil\frac{M}{Tm}\rceil$ tiles of output features are calculated. With such optimizations, the data discontinuity of features is alleviated in FP and BP. As shown in Fig.~\ref{fig:OFMchannel} (b) and Fig.~\ref{fig:IFMchannel} (b), the burst length for output features is $M^i\times Tc^i$, and the burst length for input features is $N^j\times Tc^j\_in$. Besides, the FPGA accelerator does not need to repeatedly load the input feature tiles from the DRAM. 

However, in WU, the Conv Kernel conducts MAC operations for input features and output features to calculate weight gradients, so input and output feature tiles should move in the~\textbf{Z} direction first to calculate the weight gradients of $Tm\times Tn$ weights kernels and then move in the channel direction, which is illustrated in Fig.~\ref{fig:looporig} (b). Therefore, features cannot be continuously fetched to the on-chip buffer and reused as that in the inference phase unless the on-chip memory is large enough to hold all features of each layer. When the on-chip memory cannot hold all features of a Conv layer in resource-limited FPGAs, the burst length for output features is $Tm$, and the burst length for input features is $Tn$. The data layouts are shown in Fig.~\ref{fig:OFMchannel} (c) and Fig.~\ref{fig:IFMchannel} (c).

In CNN inference, weights will not change in the whole process, and they are loaded in the same pattern for different layers. Therefore, in the inference phase, weights can be pre-allocated tile by tile to ensure continuous memory access. The pre-allocated data layout is illustrated in Fig.~\ref{fig:WEIchannel} (a). As shown in Fig.~\ref{fig:WEIchannel} (b) and Fig.~\ref{fig:WEIchannel} (d), the burst length is $M^i\times N^i$ in FP and WU.
However, as illustrated in Fig.~\ref{fig:WEIchannel} (c), the weight kernels are transposed between the input channel dimension and the output channel dimension, and the tiling scheme in BP breaks the memory access continuity. Since weights are updated after one iteration of FP, BP, and WU, it is impossible to pre-allocate them before each iteration. Therefore, data discontinuity is inevitable in BP. As shown in Fig.~\ref{fig:WEIchannel}, the burst length is $Tm$.


\subsection{Optimizing Discontinuous Memory Access}
\label{sec:Memory Allocation}

\begin{figure}[h]
  \centering
  \includegraphics[width=4.6 in
  ]{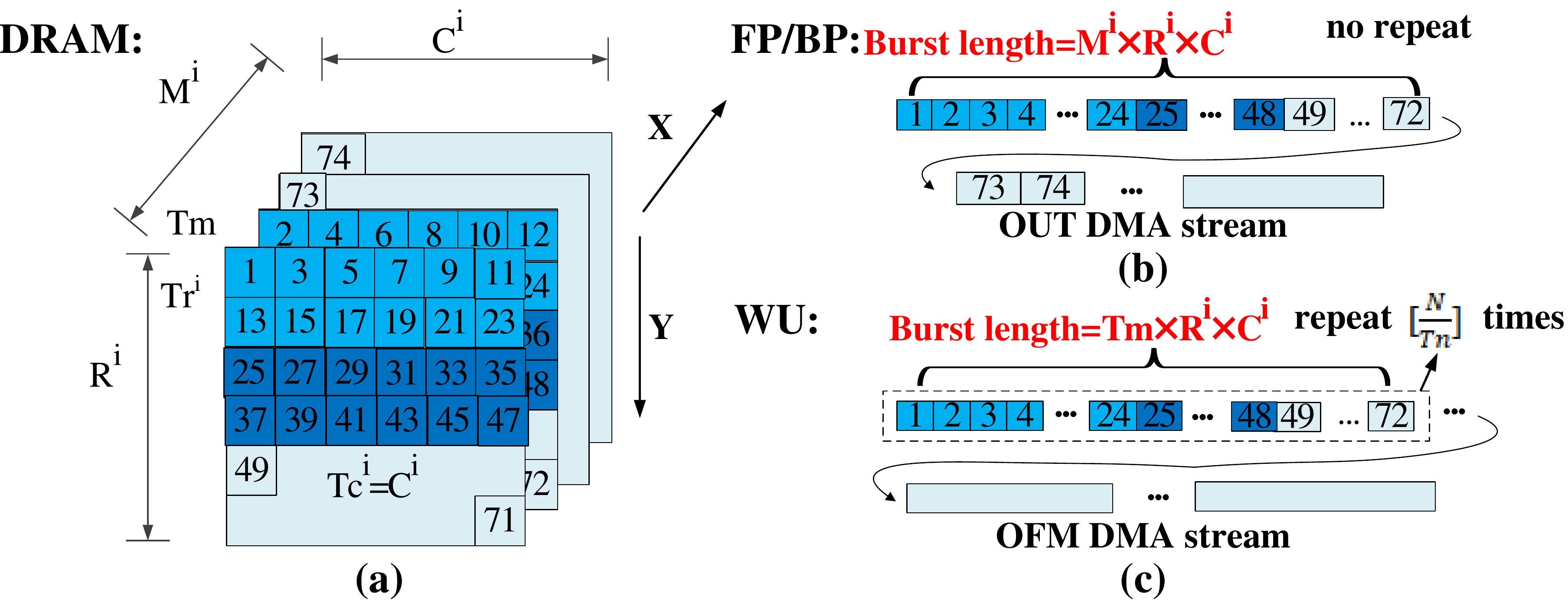}
  \caption{
  Data layout of output features after  reshaping. (a) Data stored in DRAM, (b) Data transmitted via the OUT DMA channel in FP/BP, (c) Data transmitted via the OFM DMA channel in WU.}
  \label{fig:OFMafter}
  \Description{fig:OFMafter}

  \centering
  \includegraphics[width=4.8 in
  ]{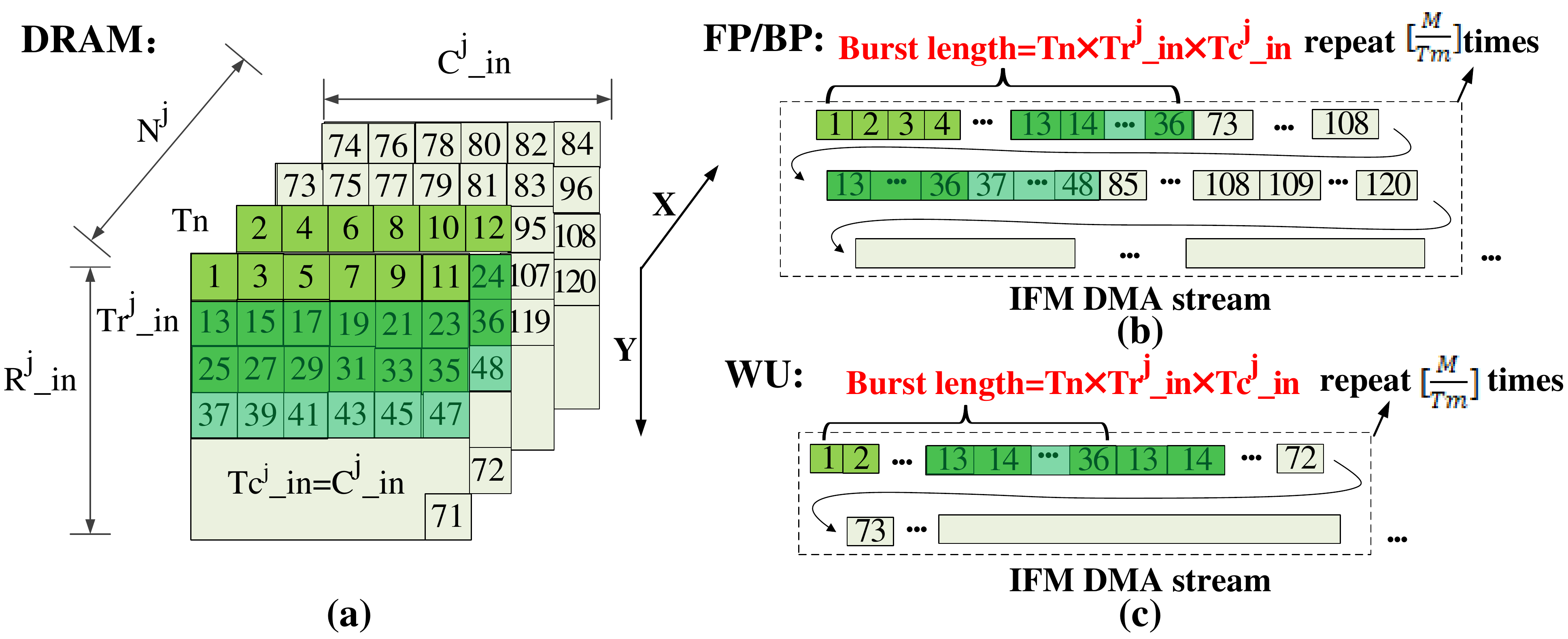}
  \caption{
  Data layout of input features after reshaping. (a) Data stored in DRAM, (b) Data transmitted via the IFM DMA channel in FP/BP, (c) Data transmitted via the IFM DMA channel in WU.}
  \label{fig:IFMafter}
  \Description{fig:IFMafter}

  \centering
  \includegraphics[width=4.8 in
  ]{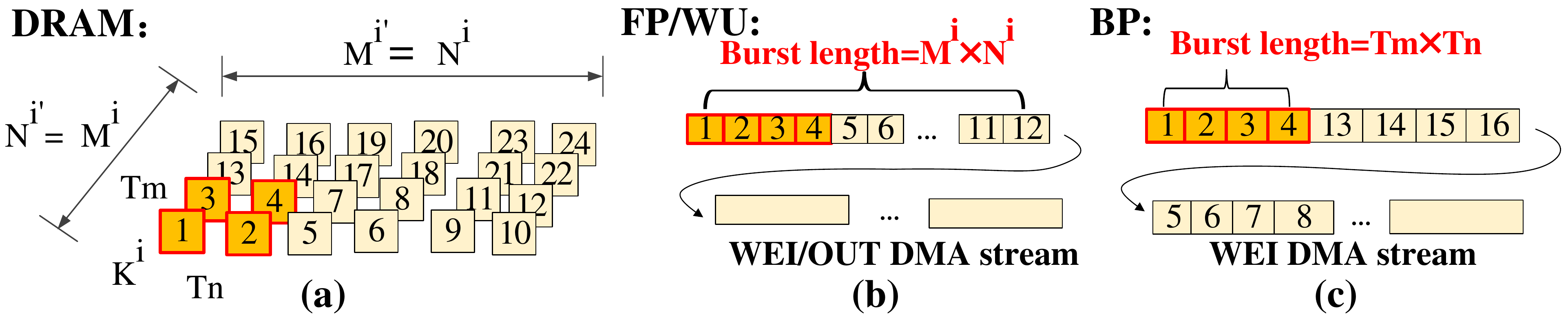}
  \caption{
  Data layout of weights after reshaping. (a) Weights stored in DRAM, (b) Weights transmitted via the WEI/OUT DMA channel in FP/WU, (c) Weights transmitted via the WEI DMA channel in BP.}
  \label{fig:WEIafter}
  \Description{fig:WEIafter}
\end{figure}

\begin{figure}[h]
  \centering
  \includegraphics[width=5.4 in
  ]{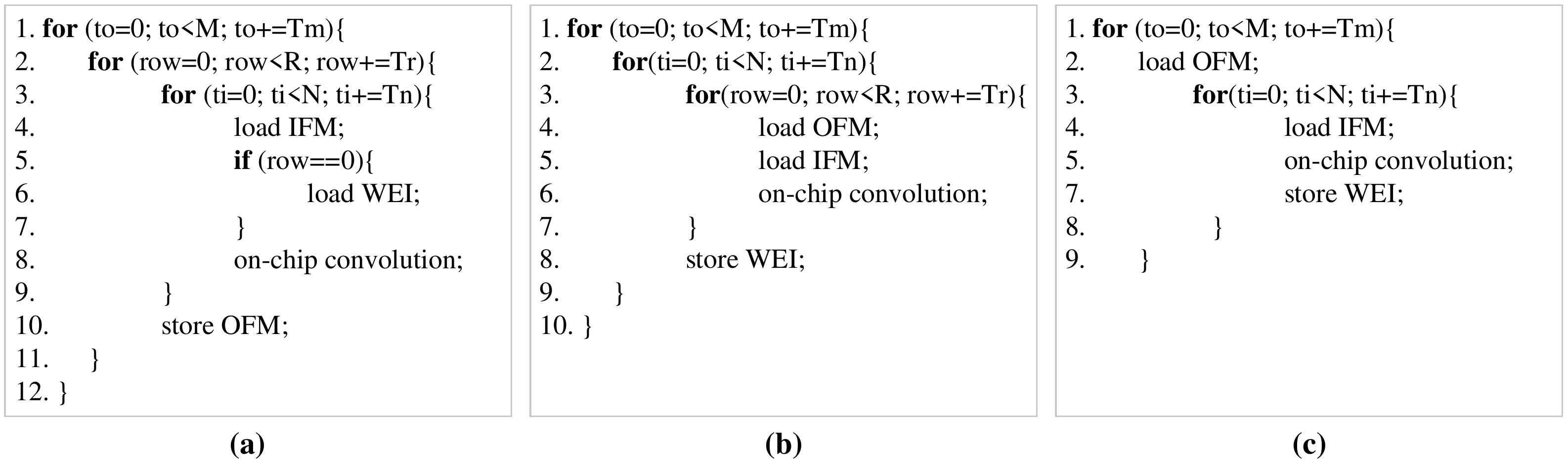}
  \caption{Pseudo-code of loop order scheduling between tiles. (a) Loop order for FP and BP, (b) Loop order for WU, (c) Loop order for WU when $R^i\leq Tr^i$.}
  \label{fig:inter}
  \Description{fig:inter}
\end{figure}

To optimize the discontinuous memory access, our data reshaping approach includes the following steps. Firstly, we achieve intra-tile continuous memory allocation for both features and weights by reorganizing the DRAM layouts for output features, input features, and weights which are shown in Fig.~\ref{fig:OFMafter}, Fig.~\ref{fig:IFMafter}, and Fig.~\ref{fig:WEIafter} respectively. Then we schedule the loop order based on the pseudo-code in Fig.~\ref{fig:inter} to achieve inter-tile continuous memory allocation. Finally, weights are reused among a mini-batch based on the proposed data layouts. {
In this section, we reorganize the data layouts and schedule the loop order together to achieve both intra-tile and inter-tile continuous memory address allocation.}

\noindent\textbf{Intra-Tile Continuous Memory Allocation:} 
Inspired by previous inference works~\cite{guan2017fp,zhang2018caffeine}, employing channel-last data layout can improve data continuity for the channel-level parallelism-based accelerator. However, as explained in Section~\ref{sec:Memory Analysis}, simply changing the data layout cannot optimize the memory access continuity in FP, BP, and WU together. The memory access patterns in the three processes need to be considered together. In CNN inference, the selection of $Tm$ and $Tn$ is flexible. However, to ensure data continuity of weights kernels in both FP and BP, we fix $Tm=Tn$ in our training accelerator so that weights can be loaded tile by tile in both FP and BP.

Fig.~\ref{fig:OFMafter} (a) shows the data layout of the output features in DRAM after data reshaping. The first $Tm$ channels of OFMs are placed in the row-column-channel pattern. The next $Tm$ channels of OFMs are followed with the same pattern. {
When applying loop tiling, we assign the tiling parameter $Tc^i=C^i$ so that data are continuous inside a tile for both FP and WU.} 
From Fig.~\ref{fig:OFMafter} (b) and (c), the burst lengths of output features during FP, BP, and WU are larger than the size of a tile.

{
The selection of $Tm=Tn$, $Tc^i=C^i$, and $Tc^i\_in=C^i\_in$ also guarantee that features of different layers share similar data layouts and tiling schemes no matter they serve as output features or input features of a Conv layer in FP/BP/WU. Therefore, the intra-tile continuity of input features is also guaranteed.} 
The data layout of the input features in DRAM for the $j$th layer is shown in Fig.~\ref{fig:IFMafter} (a). According to Fig.~\ref{fig:IFMafter} (b) and (c), the burst length equals the size of a tile.

After selecting $Tm=Tn$, weights can be placed and fetched tile by tile during FP, BP, and WU. The data layout is illustrated in Fig.~\ref{fig:WEIafter}. Before data reshaping, weights need to be repeatedly transmitted between the FPGA chip and the DRAM in FP and BP, which is inefficient especially for mini-batch training. Therefore, weight reuse based on our unique data layout is necessary which will be introduced in detail in Section~\ref{sec:reuse}. After reshaping, the burst length for FP and WU is $M^i\times N^i$, while the burst length for BP is $Tm\times Tn$, which are shown in Fig.~\ref{fig:WEIafter} (b) and (c), respectively.

\noindent\textbf{Inter-Tile Loop Order Optimization}:
{
The proposed data reshaping approach also achieves inter-tile data continuity by rescheduling the loop order in Fig.~\ref{fig:looporig}. The loop order of 1, 2, and 3 in Fig.~\ref{fig:looporig} (a) does not have data dependency. Based on our data layout, we move loop 3 to the outermost loop so that the output features share similar memory access patterns in FP/BP and WU. The loop order of off-chip data transmission in FP/BP is shown in Fig.~\ref{fig:inter} (a).} 
{
As shown in Fig.~\ref{fig:OFMafter} and Fig.~\ref{fig:IFMafter}, in FP/BP, tiles of input features are fetched in the~\textbf{X} direction first to generate the first output features tile. Then the tiles of output features are generated and stored in the~\textbf{Y} direction first, so the input features tiles movement follows the~\textbf{Y} direction as well. Then the output features tiles are generated and stored in the~\textbf{X} direction, and the access pattern of input features repeats $\lceil\frac{
M}{Tm}\rceil$ times. The burst length of output features in the OUT DMA channel is $M^i\times R^i\times C^i$. }  

The loop order in Fig.~\ref{fig:inter} (b) is adopted in WU. {
From Fig.~\ref{fig:OFMafter} and Fig.~\ref{fig:IFMafter}, tiles of both input features and output features are fetched and stored in the~\textbf{Y} direction first to calculate weights gradients for the first tile. Then weights are updated along the input channel dimension, so the input features tiles move in the~\textbf{X} direction, while the output features access pattern (the dashed box in Fig.~\ref{fig:OFMafter} (c)) repeats  $\lceil\frac{
N}{Tn}\rceil$ times. After that, weights are updated along the output channel dimension, so the output features tiles move in the~\textbf{X} direction, while the input features access pattern (the dashed box in Fig.~\ref{fig:IFMafter} (c)) repeats  $\lceil\frac{
N}{Tn}\rceil$ times.} The burst length of output features in the OFM DMA channel is $Tm\times R^i\times C^i$. When the IFM buffer and the OFM buffer are large enough to hold the $Tn\times R^i\_in\times C^i\_in$ input and $Tm^i\times R^i\times C^i$ output features, i.e. $R^i\leq Tr^i$, the output features do not need to be repeatedly loaded. The loop order can be optimized as shown in Fig.~\ref{fig:inter} (c).

\subsection{Weight Reuse in Mini-batch Training}
\label{sec:reuse}

\begin{figure}[h]
  \centering
  \includegraphics[width=5.2in
  ]{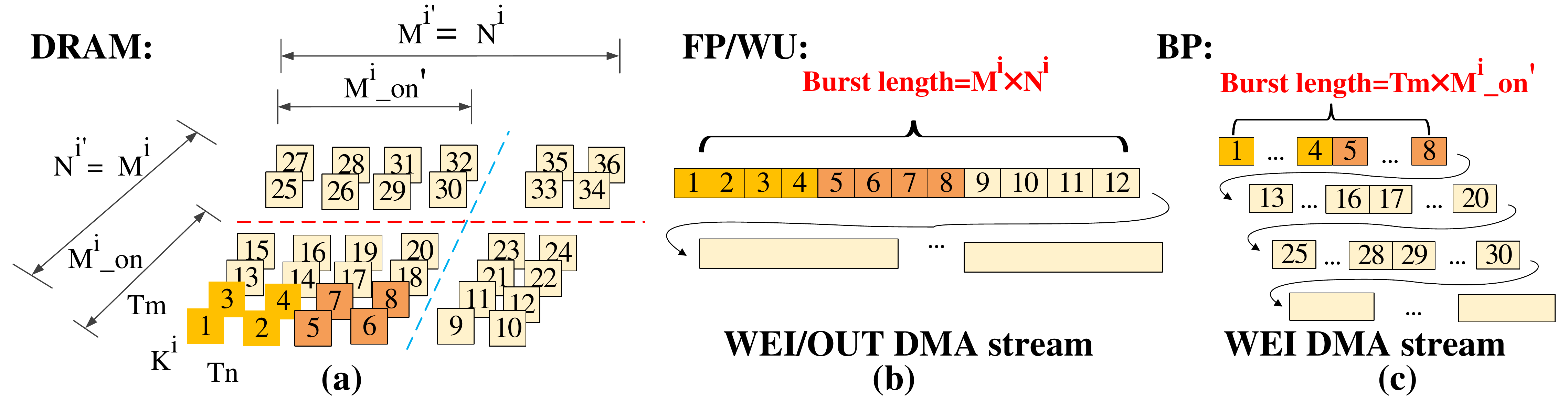}
  \caption{
  Data layout of weights after weight reuse. (a) Weights stored in DRAM, (b) Weights transmitted via the WEI/OUT DMA channel in FP/WU, (c) Weights transmitted via the WEI DMA channel in BP.}
  \label{fig:WEIreuse}
  \Description{fig:WEIreuse}
\end{figure}

{
Based on the above-mentioned optimization, we further reduce DRAM data access by reusing weights in mini-batch training.} Different from inference, training involves processing a batch of data at once, so data reuse is necessary to decrease the transmission times of weights between on-chip buffer and off-chip memory. On FPGAs, a BRAM bank size is large enough to store multiple tiles of weights. Therefore, we propose a weight reuse strategy based on our data layout. Thanks to our loop order shown in Fig.~\ref{fig:inter}, we can load weights only when the accelerator processes the output feature tile lying in the first row. As illustrated in Fig.~\ref{fig:WEIreuse}, when the accelerator processes a tile of features in the first row of the first image in a batch, $M^i\_{on}\times N^i\times K^i\times K^i$ weights are loaded and stored in the WEI double buffers, where $M^i\_{on}$ is the multiple of $Tm$ depending on the on-chip BRAM resources. After the first $M^i\_{on}$ channels of OFMs in the image are processed, the first $M^i\_{on}$ channels of OFMs of the next image will be processed, so weights do not need to be uploaded again. The next $M^i\_{on}$ channels of the first image will be processed after the first $M^i\_{on}$ channels of all images in the batch are processed. Therefore, weights do not need to be transmitted back and forth. After the above-mentioned steps, the burst length is $M^i\times N^i$ for FP/WU and $Tm\times M^i\_{on}'$ for BP, which are illustrated in Fig.~\ref{fig:WEIreuse} (b) and (c) respectively.

\begin{figure}[h]
  \centering
  \includegraphics[width=4.8 in
  ]{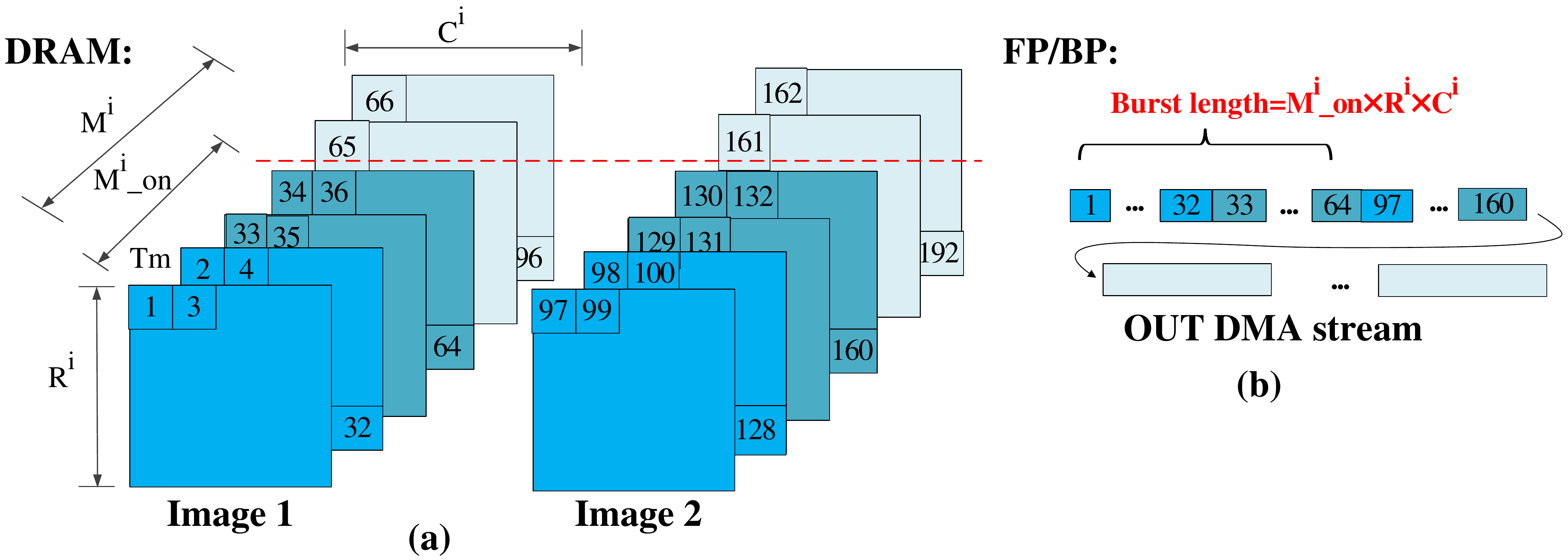}
  \caption{
  Data layout of output features in mini-batch training after weight reuse. (a) Output features of two images in a batch, (b) Output features transmitted the via the OUT DMA channel in FP/BP.}
  \label{fig:reusetradeoff}
  \Description{fig:reusetradeoff}
\end{figure}

In mini-batch training, weight reuse will not affect the burst length of output features in WU and input features. For output features in FP and BP, after the first $M^i\_on$ channels of OFMs of the first image are transmitted to DRAM, the next image of the batch will be processed before other channels of the prior image. Therefore, as shown in Fig.~\ref{fig:reusetradeoff}, the burst length is $M^i\_on\times R^i\times C^i$.

\section{Performance and Resource Model}
\label{sec:model}
In this section, we establish an analytic model to calculate the latency and resources for our design. Unlike previous works~\cite{JasonOptimizing,jiang2019achieving} which only focused on the performance of a bare accelerator running on separate Conv layers, our model considers the discontinuity of off-chip memory access in a realistic end-to-end training process. Based on the model, we build a scheduling tool to determine design parameters for given FPGA devices and given network models. 

\subsection{Performance Model}
\label{sec:timing}

For a $[M^i,N^i,R^i,C^i,K^i,K^i,S^i]$ Conv layer $i$, we assume the parameters of a tile is $[Tm,Tn,Tr^i,Tc^i]$. $Tm$ and $Tn$ are fixed for all layers since they are determined by the number of DSPs, while $Tr^i$ and $Tc^i$ are adjustable according to different layer parameters. In our design, $Tm=Tn$, and $Tc^i=C^i$. The computation latency of a tile of features in FP, BP, and WU can be represented as $t^i_{COMP}=Tr^i\times Tc^i\times K^i\times K^i$ clock cycles.

The continuity of memory address significantly impacts the off-chip communication efficiency. To consider the memory access discontinuity, we assume the start time of the DMA stream is $t_{start}$. When discontinuity happens, DMA restarts. We have tested the start time on both the PYNQ-Z1 and the ZCU102 board, and $t_{start}\approx 400$ cycles under 100MHz clock. We determine the data width parameters $p$ to model the off-chip/on-chip communication bandwidth. For 32-bit floating-point, if the DMA stream width is 128 bits, $p=4$. Since the burst length of input features equals the size of a tile, discontinuity happens every time a tile of input features are fetched. The latency of loading a tile of input features is formulated as $t^i_{IFM}=t_{start}+\lceil\frac{Tn}{p}\rceil \times ((Tr^i-1)\times S^i+K^i)\times ((Tc^i-1)\times S^i+K^i)$ clock cycles. The weights loading latency can be represented as $t^i_{WEI}=\lceil\frac{Tm\times Tn}{p}\rceil\times K^i\times K^i$ clock cycles, and the latency of storing a tile of output features is formulated as $t^i_{OUT}=\lceil\frac{Tm}{p}\rceil\times Tr^i\times Tc^i$ clock cycles. $t_{start}$ is added to $t^i_{WEI}$ and $t^i_{OUT}$ only when the discontinuity happens, which will be discussed in detail as follows. We define $t^i_{LOAD}=\max\left\{t^i_{IFM},t^i_{WEI}\right\}$, $t^i_{PROD1}=\max\left\{t^i_{IFM},t^i_{COMP}\right\}$, $t^i_{PROD2}=\max\left\{t^i_{LOAD},t^i_{COMP}\right\}$, and $t^i_{STORE}=\max\left\{t^i_{COMP},t^i_{OUT}\right\}$.

We assume in layer $i$, $M^i\_on\times Tn\times K^i \times K^i$ weights  are stored on-chip. If the batch size is $B$, weights will be loaded only during the iteration when the proposed accelerator processes the first image in the batch. For other iterations, the latency of processing $M^i\_on$ channels of an image in FP can be formulated as follows.
\begin{equation}
    Lat1^i=\lceil\frac{N^i}{Tn}-1\rceil \times t^i_{PROD1}+t^i_{IFM}+t^i_{COMP} 
\label{Lat1}
\end{equation}
\begin{equation}
    Lat2^i=\lceil\frac{N^i}{Tn}-1\rceil \times t^i_{PROD1}+t^i_{IFM}+t^i_{STORE} 
\label{Lat2}
\end{equation}
\begin{equation}
    Lat3^i=(\lceil\frac{M^i\_on}{Tm}\rceil \times\lceil\frac{R^i}{Tr^i}\rceil-1)\times Lat2^i+Lat1^i+t^i_{OUT}+t^i_{start} 
\label{Lat3}
\end{equation}

Weights need to be loaded when our accelerator processes the first image in the mini-batch. In FP, $t_{start}$ can be neglected in weight transmission since the burst length equals the size of weights, which means the addresses are continuous during the whole Conv layer. Therefore, the latency of the proposed accelerator processing $M^i\_on$ channels of the first image can be formulated as follows.
\begin{equation}
    Latb1^i=\lceil\frac{N^i}{Tn}-1\rceil\times t^i_{PROD2}+t^i_{LOAD}+t^i_{COMP} 
\label{Latb1}
\end{equation}
\begin{equation}
    Latb2^i=\lceil\frac{N^i}{Tn}-1\rceil\times t^i_{PROD2}+t^i_{LOAD}+t^i_{STORE} 
\label{Latb2}
\end{equation}
\begin{equation}
    Latb3^i=\lceil\frac{M^i\_on}{Tm}\rceil \times\lceil\frac{R^i}{Tr^i}-1\rceil\times Lat2^i+\lceil\frac{M^i\_on}{Tm}-1\rceil\times Latb2^i+Latb1^i+t^i_{OUT}+t^i_{start} 
\label{Latb3}
\end{equation}

The latency of our accelerator processing the whole Conv layer in FP is formulated in Eq.~\eqref{LatFP}.
\begin{equation}
    Lat^i=\lceil\frac{M^i}{M^i\_on}\rceil\times((B-1)\times Lat3^i+Latb3^i).
\label{LatFP}
\end{equation}

In BP, the situation is similar to that in FP, except that the addresses of weights are discontinuous after $M^i\_on$ channels. The accelerator loads $M^i\_on\times Tn\times K^i\times K^i$ weights together when processing the first tile of the first image, so it costs $t^i_{WEI}=\lceil\frac{M^i\_on\times Tn}{p}\rceil\times K^i\times K^i+t^i_{start}$ clock cycles. $Lat1^i$, $Lat2^i$, $Lat3^i$, $Lat2^i$, $Latb1^i$, and $Latb2^i$ remain unchanged, while $Latb3^i=(\lceil\frac{M^i\_on}{Tm}\rceil\times\lceil\frac{R^i}{Tr^i}\rceil-1)\times Lat2^i+Latb1^i+t^i_{OUT}+t^i_{start}$.

In WU, loss features are loaded from the off-chip memory to the OFM buffer. Transmitting a tile of loss features costs $t^i_{OFM}=t^i_{start}+Tr^i\times Tc^i\times\lceil\frac{Tm}{p}\rceil$ clock cycles. Weights are updated after all the gradients of the batch are accumulated, so transmitting the updated weights costs the same time as loading weights, which means $t^i_{OUT}\!=\!t^i_{WEI}$. Same with FP, $t_{start}$ can be neglected when calculating $t^i_{WEI}$. We define $t^i_{LOAD}=\max\left\{t^i_{IFM},t^i_{OFM}\right\}$, $t^i_{PROD1}=\max\left\{t^i_{LOAD},t^i_{COMP}\right\}$, $t^i_{PROD2}=\max\left\{t^i_{IFM},t^i_{COMP}\right\}$, and $t^i_{STORE}=\max\left\{t^i_{COMP},t^i_{OUT}\right\}$. The latency of WU of the $i$th Conv layer is formulated as follows.
\begin{equation}
    Lat1^i=\lceil\frac{R^i}{Tr^i}-1\rceil\times t^i_{PROD1}+t^i_{LOAD}+t^i_{COMP} 
\label{Lat1WU}
\end{equation}
\begin{equation}
    Latb1^i=\lceil\frac{R^i}{Tr^i}-1\rceil\times t^i_{PROD1}+t^i_{LOAD}+t^i_{STORE} 
\label{Latb1WU}
\end{equation}
\begin{equation}
    Lat^i=(((B-1)\times\lceil\frac{M^i\_on}{Tm}\rceil\times\lceil\frac{N^i}{Tn}\rceil+1)\times Lat1^i+(\lceil\frac{M^i\_on}{Tm}\rceil\times\lceil\frac{N^i}{Tn}\rceil-1)\times Latb1^i+t^i_{OUT})\times\lceil\frac{M^i}{M^i\_on}\rceil
\label{LatbWU}
\end{equation}

As illustrated in Fig.~\ref{fig:inter} (c), when $R^i\leq Tr^i$, the output features do not need to be repeatedly loaded. Under this circumstance, the latency of WU is formulated as follows.
\begin{equation}
    Lat1^i=\lceil\frac{N^i}{Tn^i}-1\rceil\times t^i_{PROD2}+t^i_{LOAD}+t^i_{COMP} 
\label{Lat1WUc}
\end{equation}
\begin{equation}
    Latb1^i=\lceil\frac{N^i}{Tn^i}-1\rceil\times (t^i_{PROD2}+t^i_{OUT})+t^i_{LOAD}+t^i_{COMP}+t^i_{OUT}
\label{Latb1WUc}
\end{equation}
\begin{equation}
    Lat^i=\lceil\frac{M^i}{M^i\_on}\rceil\times\lceil\frac{M^i\_on}{Tm}\rceil\times((B-1)\times Lat1^i+Lat1b^i)
\label{LatbWUc}
\end{equation}

\subsection{Resource Model}
\label{sec:resource}

For Conv layers, the on-chip resources that need to be considered for Conv layers include DSPs and BRAMs. For DSPs, $Tm\times Tn$ MAC operations are conducted in parallel. Therefore, the computation constraint is shown in Eq.~\eqref{eq:DSP}, where $q$ is the factor depending on data types. On Xilinx FPGAs, each MAC utilizes 5 DSPs for 32-bit floating-point, so $q=5$ in the proposed design. In terms of on-chip memory, we select double buffers to load and store data and conduct Conv operations in parallel. The number of BRAM banks for each IFM buffer and OFM buffer are shown in Eq.~\eqref{eq:IFM} and Eq.~\eqref{eq:OFM} respectively. The notation BITs is the data bit-width adopted in the design. For the Weight buffer, we place $M^i\_on\!\times\!N^i$ kernels together for weight reuse. These data are scattered in double buffers. The number of BRAM banks for one Weight buffer is shown in Eq.~\eqref{eq:WEI}. The on-chip memory constraint is shown in Eq.~\eqref{eq:BRAM}.
\begin{equation}
    D_{Conv}=q\times Tm \times Tn< total\ DSPs\ number 
\label{eq:DSP}
\end{equation}
\begin{equation}
    B_{IFM}=\max\limits_i B^i_{IFM}=\max\limits_i\left\{Tn\times\lceil\frac{((Tr^i-1)\times S^i+K^i)\times ((Tc^i-1)\times S^i+K^i)\times BITs}{Size\ of\ a\ BRAM\ Bank}\rceil\right\}
\label{eq:IFM}
\end{equation}
\begin{equation}
    B_{OFM}=\max\limits_i B^i_{OFM}=\max\limits_i\left\{Tm\times\lceil\frac{Tr^i\times Tc^i\times BITs}{Size\ of\ a\ BRAM\ Bank}\rceil\right\} 
\label{eq:OFM}
\end{equation}
\begin{equation}
    B_{WEI}=\max\limits_i B^i_{WEI}=\max\limits_i\left\{Tm\times Tn\times\lceil\frac{K^i\times K^i\times\lceil\frac{N^i}{2\times Tn}\rceil\times\lceil\frac{M^i\_on}{Tm}\rceil\times BITs}{Size\ of\ a\ BRAM\ Bank}\rceil\right\} 
\label{eq:WEI}
\end{equation}
\begin{equation}
    B_{Conv}=2\times(B_{IFM}+B_{OFM}+B_{WEI})<total\ BRAMs\ number 
\label{eq:BRAM}
\end{equation}

It should be noted that in realistic end-to-end system design, the boundary of $D_{Conv}$ and $B_{Conv}$ should be slightly smaller than the total DSPs and BRAMs numbers. It is because except for the MAC operations, several operations also take up a small fraction of on-chip resources. For example, some non-Conv layers (e.g. maximum pooling, average pooling, ReLU, etc.), which are inevitable in practical end-to-end training processes need extra DSPs to make comparisons and extra BRAMs to buffer the indexes. Besides, some neural networks have irregular weights kernel shapes for different Conv layers. Adding an extra buffer to fetch a tile of weights from the on-chip Weight buffer to the Conv Kernel can relieve the routing congestion in realistic FPGA implementation. Besides, since FP, BP, WU have different loop orders, extra DSPs are utilized to calculate BRAM addresses under different layer parameters. This address calculation is much more complex than that in inference. Therefore, in practical design, the estimated boundary of the on-chip resources should be set lower than the available resources empirically. The details will be further explained in Section~\ref{sec:schedule}.

\subsection{Computation and Memory Resources Scheduling Tool}
\label{sec:schedule}

Based on the above-mentioned model, we build a computation and memory resources scheduling tool for different devices and different networks. Algorithm~\ref{alg:schedule} shows the framework of our scheduling tool. As mentioned in Section~\ref{sec:resource}, $D_{Conv}$ and $B_{Conv}$ are lower than the total DSPs and BRAMs numbers in realistic FPGA implementation. Therefore, it is wise to set a boundary for $D_{Conv}$ and $B_{Conv}$ that is lower than the available on-chip resources. According to the experimental results in Section~\ref{sec:experiment}, assigning $80\%$ of DSPs and $75\%$ BRAMs to the estimated boundary for $D_{Conv}$ and $B_{Conv}$ should be enough.  Then we determine $Tm$ and $Tn$ according to the DSPs number. Then we choose the optimal $Tr^i$, $Tc^i$, and $M^i\_{on}$ for each layer according to Eq.~\eqref{Lat1} - \eqref{LatbWUc}. Specifically, in steps~\ref{inf:BIFMBOFM1} and ~\ref{inf:BIFMBOFM2}, we find the lower bound for $B_{IFM}$ and $B_{OFM}$ by assuming that the buffers can only hold one row for the largest feature maps. Then, from step~\ref{BWEI1} to step~\ref{BWEI2}, we try to assign resources for Weight buffers so that they can hold as many weights for each layer as possible. After we determine $B_{WEI}$ and $M^i\_{on}$ for each layer, we re-assign IFM and OFM buffers under the constraints shown in Eq.~\eqref{eq:IFM},~\eqref{eq:OFM}, and~\eqref{eq:BRAM}, and find the optimal $Tr^i$ and $Tc^i$ for each layer.   After $Tm$, $Tn$, and $[Tr^i, Tc^i, M^i\_{on}]_{1\le i\le n}$ are determined, we can calculate the DMA start addresses for each layer off-line based on the data reshaping approach in Section~\ref{sec:memory}. 

\begin{algorithm}[htb] 
\caption{Computation and Memory Resources Scheduling} 
\label{alg:schedule} 
\begin{algorithmic}[1]
\REQUIRE ~~\\ 
CNN layers parameters $[M^i, N^i, R^i, C^i, K^i, K^i, S^i]_{1\le i\le n}$, batch size $B$, data type parameter $q$, DMA stream width, total DSPs number, total BRAMs number;
\ENSURE ~~\\ 
$Tm$, $Tn$, $[Tr^i, Tc^i, M^i\_{on}]_{1\le i\le n}$, $B_{IFM}$, $B_{OFM}$, $B_{WEI}$;
\STATE Estimate the boundary for $D_{Conv}$ and $B_{Conv}$; 
\STATE Assign $Tm$, $Tn$ according to Eq.~\eqref{eq:DSP}, while $Tm=Tn$; 
\STATE Find $k=\mathop{\arg\max}\limits_{i}\left\{R^i\times C^i\right\}$;\label{inf:BIFMBOFM1}
\STATE Determine the lower bound for $B_{IFM}$ and $B_{OFM}$, i.e.  $\inf B_{IFM}=B^k_{IFM}$, $\inf B_{OFM}=B^k_{OFM}$, when $Tc^k=C^k$, $Tr^k=1$;\label{inf:BIFMBOFM2}
\FOR{$i=1$; $i\le n$; $i++$ }\label{BWEI1}
\STATE Calculate $B^i_{WEI}$ based on ~\eqref{eq:WEI} when $M^i\_{on}=M^i$, and initialize $l=1$;
\IF{$2\times (\inf B_{IFM}+\inf B_{OFM}+B^i_{WEI})\geq estimated\ B_{Conv}\ boundary$}
\label{if:BWEI}
\STATE l++;
\STATE Find the minimal $M^i\_{on}$ satisfying $\frac{M^i}{l}\le M^i\_{on}$, $M^i\_{on}\mod Tm =0$, and go to step~\ref{if:BWEI};
\ENDIF
\ENDFOR
\STATE Calculate $B_{WEI}$ and $M^i\_{on}$ for each layer based on Eq.~\eqref{eq:WEI};\label{BWEI2}
\FOR{$i=1$; $i\le n$; $i++$ }
\STATE Set $Tc^i=C^i$, and select all $Tr^i_m$ satisfying Eq.~\eqref{eq:IFM},~\eqref{eq:OFM}, and~\eqref{eq:BRAM}, where $1\le Tr^i_m\le R^i$;
\STATE Determine $Tr^i=\mathop{\arg\min}\limits_{m}Lat^i_m$ based on Eq.~\eqref{Lat1}-~\eqref{LatbWUc};
\ENDFOR
\STATE Calculate $B_{IFM}$ and $B_{OFM}$ based on Eq.~\eqref{eq:IFM},~\eqref{eq:OFM};
\end{algorithmic}
\end{algorithm}

\section{Experiments}
\label{sec:experiment}

The proposed work is evaluated on edge-level FPGAs PYNQ-Z1 and ZCU102 with working frequency at 100MHz. The accelerator is designed with Vivado HLS, which generates IP core from C language. The obtained IP cores are connected, synthesized, and implemented in Vivado (v2019.1). The Vivado Project Summary reports resource utilization and power after implementation. Finally, we employ Xilinx SDK to
program SoC on PYNQ-Z1 and ZCU102 to achieve end-to-end CNN training.

\subsection{Effectiveness of The Data Reshaping Approach}
\label{sec:effectiveness approach}

In this section, we need to validate the effectiveness of the proposed data reshaping approach ($[Tm, Tn]=[16, 16]$). We test the Conv layers of the AlexNet on ZCU-102. We select the batch size $B$ as 4 and the DMA stream width as 128 bits. {
We adopt the results using the BCHW data layout and the results using the BHWC data layout as baselines ($[Tm_{Base}, Tn_{Base}]=[32, 8]$). The BCHW pattern does not involve any optimization. For the BHWC pattern, $N/Tn$ tiles of input features and $M/Tm$ tiles of the output features are buffered in the on-chip BRAM for data reuse based on the loop order in the inference phase. Weights are pre-allocated tile by tile based on the data flow in inference. {
The comparisons are shown in Tables~\ref{tab:effectiveness approach baseline},~\ref{tab:effectiveness approach baselinechannel}, and~\ref{tab:effectiveness approach}}.}

\begin{table}
  \caption{
  Experimental Results of The Baseline with The BCHW Data Layout}
  \label{tab:effectiveness approach baseline}
  \begin{tabular}{cccccc}
    \toprule
    \multirow{2}{*}{AlexNet}
    &\multirow{2}{*}{Process}
    &\multirow{2}{*}{$[Tr^i_{Base}, Tc^i_{Base}]$} &\multirowcell{2}{Acceleration\\(cycles)}  &\multirowcell{2}{Reallocation\\(cycles)} &\multirowcell{2}{Total\\(cycles)}\\\\
    \midrule
    Conv 1
    &\multirowcell{3}{FP\\BP\\WU} &\multirowcell{3}{$[11, 11]$\\N/A\\$[11,11]$} &\multirowcell{3}{6,732,837\\N/A\\4,496,029}  &\multirowcell{3}{151,846,336\\N/A\\152,110,235}
    &\multirowcell{3}{158,579,173\\N/A\\156,606,264}
    \\\\\\  \hline
    Conv 2
    &\multirowcell{3}{FP\\BP\\WU} &\multirowcell{3}{$[27, 27]$\\$[27, 27]$\\$[27, 27]$} &\multirowcell{3}{7,105,292\\7,066,705\\9,258,823}  &\multirowcell{3}{69,743,160\\68,271,764\\57,303,397}
    &\multirowcell{3}{76,848,452\\75,338,469\\66,562,220}
    \\\\\\  \hline
    Conv 3
    &\multirowcell{3}{FP\\BP\\WU} &\multirowcell{3}{$[13, 13]$\\$[13, 13]$\\$[13, 13]$} &\multirowcell{3}{2,410,532\\2,401,320\\4,448,898}  &\multirowcell{3}{101,062,954\\98,646,892\\83,566,193}
    &\multirowcell{3}{103,473,486\\101,048,212\\88,015,091}
    \\\\\\  \hline
    Conv 4
    &\multirowcell{3}{FP\\BP\\WU} &\multirowcell{3}{$[13, 13]$\\$[13, 13]$\\$[13, 13]$} &\multirowcell{3}{3,596,425\\3,596,400\\6,669,238}  &\multirowcell{3}{150,012,382\\149,621,995\\126,214,297}
    &\multirowcell{3}{153,608,807\\153,218,395\\132,883,535}
    \\\\\\  \hline
    Conv 5
    &\multirowcell{3}{FP\\BP\\WU} &\multirowcell{3}{$[13, 13]$\\$[13, 13]$\\$[13, 13]$} &\multirowcell{3}{2,401,212\\2,410,637\\4,448,751}  &\multirowcell{3}{102,632,162\\99,408,011\\84,518,969}
    &\multirowcell{3}{105,033,374\\101,818,648\\88,967,720}
    \\\\\\  \hline
    \textbf{Total}
    & & &\textbf{67,043,099}
    &\textbf{1,494,958,747}
    &\textbf{1,562,001,846}
    \\
  \bottomrule
\end{tabular}
\end{table}

\begin{table}
  \caption{
  Experimental Results of The Baseline with The BHWC Data Layout and Data Reuse}
  \label{tab:effectiveness approach baselinechannel}
  \begin{tabular}{cccccc}
    \toprule
    \multirow{2}{*}{AlexNet}
    &\multirow{2}{*}{Process}
    &\multirow{2}{*}{$[Tr^i_{Base}, Tc^i_{Base}]$} &\multirowcell{2}{Acceleration\\(cycles)}  &\multirowcell{2}{Reallocation\\(cycles)} &\multirowcell{2}{Total\\(cycles)}\\\\
    \midrule
    Conv 1
    &\multirowcell{3}{FP\\BP\\WU} &\multirowcell{3}{$[11, 11]$\\N/A\\$[11,11]$} &\multirowcell{3}{8,094,251\\N/A\\4,495,794}  &\multirowcell{3}{N/A\\N/A\\161,048,775}
    &\multirowcell{3}{8,094,251\\N/A\\165,544,569}
    \\\\\\  \hline
    Conv 2
    &\multirowcell{3}{FP\\BP\\WU} &\multirowcell{3}{$[27, 27]$\\$[27, 27]$\\$[27, 27]$} &\multirowcell{3}{7,383,996\\7,382,504\\7,848,249}  &\multirowcell{3}{N/A\\68,200,715\\N/A}
    &\multirowcell{3}{7,383,996\\75,583,219\\7,848,249}
    \\\\\\  \hline
    Conv 3
    &\multirowcell{3}{FP\\BP\\WU} &\multirowcell{3}{$[13, 13]$\\$[13, 13]$\\$[13, 13]$} &\multirowcell{3}{2,531,247\\2,529,216\\3,345,845}  &\multirowcell{3}{N/A\\100,372,954\\N/A}
    &\multirowcell{3}{2,531,247\\102,902,170\\3,345,845}
    \\\\\\  \hline
    Conv 4
    &\multirowcell{3}{FP\\BP\\WU} &\multirowcell{3}{$[13, 13]$\\$[13, 13]$\\$[13, 13]$} &\multirowcell{3}{3,745,972\\3,745,922\\4,999,576}  &\multirowcell{3}{N/A\\148,657,460\\N/A}
    &\multirowcell{3}{3,745,972\\152,403,382\\4,999,576}
    \\\\\\  \hline
    Conv 5
    &\multirowcell{3}{FP\\BP\\WU} &\multirowcell{3}{$[13, 13]$\\$[13, 13]$\\$[13, 13]$} &\multirowcell{3}{2,529,173\\2,531,318\\3,364,408}  &\multirowcell{3}{N/A\\100,586,051\\N/A}
    &\multirowcell{3}{2,529,173\\103,117,369\\3,364,408}
    \\\\\\  \hline
    \textbf{Total}
    & & &\textbf{64,527,471}
    &\textbf{578,865,955}
    &\textbf{643,393,426}
    \\
  \bottomrule
\end{tabular}
\end{table}

\begin{table}
  \caption{
  Experimental Results Validating Data Reshaping Approach}
  \label{tab:effectiveness approach}
  \begin{tabular}{ccccc}
    \toprule
    \multirow{2}{*}{AlexNet}
    &\multirow{2}{*}{Process}
    &\multirow{2}{*}{$[Tr^i, Tc^i]$}
    &\multirowcell{2}{Without Weight Reuse\\(cycles)}  &\multirowcell{2}{After Weight Reuse\\(cycles)}
    \\\\
    \midrule
    Conv 1
    &\multirowcell{3}{FP\\BP\\WU} &\multirowcell{3}{$[2, 55]$\\N/A\\$[2,55]$} 
    &\multirowcell{3}{11,498,545\\N/A\\9,598,744} &\multirowcell{3}{11,419,835\\N/A\\9,299,086}
    \\\\\\  \hline
    Conv 2
    &\multirowcell{3}{FP\\BP\\WU} &\multirowcell{3}{$[27, 27]$\\$[27, 27]$\\$[27, 27]$} &\multirowcell{3}{7,283,187\\7,128,663\\7,910,148}   &\multirowcell{3}{7,312,794\\7,146,578\\7,430,533}
    \\\\\\  \hline
    Conv 3
    &\multirowcell{3}{FP\\BP\\WU} &\multirowcell{3}{$[13, 13]$\\$[13, 13]$\\$[13, 13]$}  &\multirowcell{3}{2,491,672\\2,461,694\\3,402,418}
     &\multirowcell{3}{2,510,310\\2,671,392\\2,706,696}
    \\\\\\  \hline
    Conv 4
    &\multirowcell{3}{FP\\BP\\WU} &\multirowcell{3}{$[13, 13]$\\$[13, 13]$\\$[13, 13]$}   &\multirowcell{3}{3,689,930\\3,688,961\\5,053,485}
    &\multirowcell{3}{3,708,934\\3,972,757\\4,014,651}
    \\\\\\  \hline
    Conv 5
    &\multirowcell{3}{FP\\BP\\WU} &\multirowcell{3}{$[13, 13]$\\$[13, 13]$\\$[13, 13]$}   &\multirowcell{3}{2,462,778\\2,490,897\\3,373,373}
     &\multirowcell{3}{2,475,263\\2,686,910\\2,677,726}
    \\\\\\  \hline
    \textbf{Total}
    & & &\textbf{72,534,495}
    &\textbf{70,033,465}
    \\
  \bottomrule
\end{tabular}
\end{table}

{
As mentioned in Section~\ref{sec:Motivations}, our goal is to design a general accelerator supporting end-to-end training with both dense and small networks without sacrificing precision, so it is necessary to appropriately manage external memory access and allocate on-chip buffers. When applying loop tiling, the tiling schemes involved in the accelerator design break the continuity of data addresses in DRAM and thus reduce the transmission efficiency between on-chip buffer and off-chip DRAM. Table~\ref{tab:effectiveness approach baseline} shows the experimental results of our baseline which is a bare accelerator with the unified channel-level parallelism-based convolution kernel. It does not involve any optimizations related to the off-chip DRAM access policy.} As illustrated in Section~\ref{sec:memory}, the burst length before data reshaping is much smaller than the size of a tile. To ensure that the accelerator conducts MAC operations with correct features and weights matrices in realistic end-to-end training, data should be reallocated before being transmitted from DRAM to the on-chip accelerator. Therefore, our baseline includes the on-chip acceleration time and off-chip reallocation time. After applying data reshaping, data can be fetched from DRAM to the accelerator directly without extra reallocation. 

For Conv 1, the number of input channels is only 3, $Tn_{Base}=8$, and $Tn=16$. Therefore, $5/8$ computation resources remain idle for the baseline, while $13/16$ computation resources for our proposed design remain idle. That's why the acceleration time for the baseline is shorter than the latency in our proposed design. However, features should be reallocated before entering the next layer (for FP) or after being generated from the prior layer (for WU). As shown in Table~\ref{tab:effectiveness approach baseline}, the reallocation time is much longer than the acceleration time. For Conv 2 to Conv 5, $Tr^i\geq R^i$ and $Tc^i\geq C^i$, so features do not need to be reallocated between adjacent layers, but weights still need to be reallocated before entering the Conv layer (for FP and BP) or updated from the Conv layer (for WU). To sum up, the total acceleration time for the baseline is close to that for our proposed design under the same degree of parallelism ($Tm_{Base}\times Tn_{Base}=Tm\times Tn$) and tile size boundary of features  ($\max\limits_i Tr^i_{Base}\times \max\limits_i Tc^i_{Base}=\max\limits_i Tr^i\times \max\limits_i Tc^i$), but the extra reallocation time in realistic end-to-end training is even longer than the acceleration time. Therefore, accelerating without considering the actual data layout in DRAM between adjacent layers is inefficient in realistic end-to-end training. 

{
The baseline in Table~\ref{tab:effectiveness approach baselinechannel} uses the BHWC data layout and applies data reuse to alleviate the discontinuous memory access. As illustrated in Figs.~\ref{fig:OFMchannel}-\ref{fig:WEIchannel}, features and weights are continuous in a long burst length in FP, so data are not reallocated during the Conv layers. Such an approach is efficient in the inference phase. In BP, although the memory access pattern of features is the same as that in FP, the weights transmission patterns are quite different. As shown in Fig.~\ref{fig:WEIchannel} (c), the burst length is much less than the size of a tile, so weights should be reallocated in each Conv layer. The extra reallocation time is much longer than the acceleration time. In WU, the on-chip buffer can hold all the features for Conv 2-Conv 5 layers, so it is practical to load all the features to the FPGA chip without extra reallocation. However, in the Conv 1 layer, the on-chip memory cannot hold all the features. Even though the input features can be pre-allocated before entering into the neural network since they serve as the inputs for the whole process, the output features which are calculated in BP cannot be allocated ahead of time. Therefore, the Conv 1 layer also requires extra reallocation time in WU, which is quite inefficient.
}

\begin{figure}[h]
  \centering
  \includegraphics[width=3.8 in
  ]{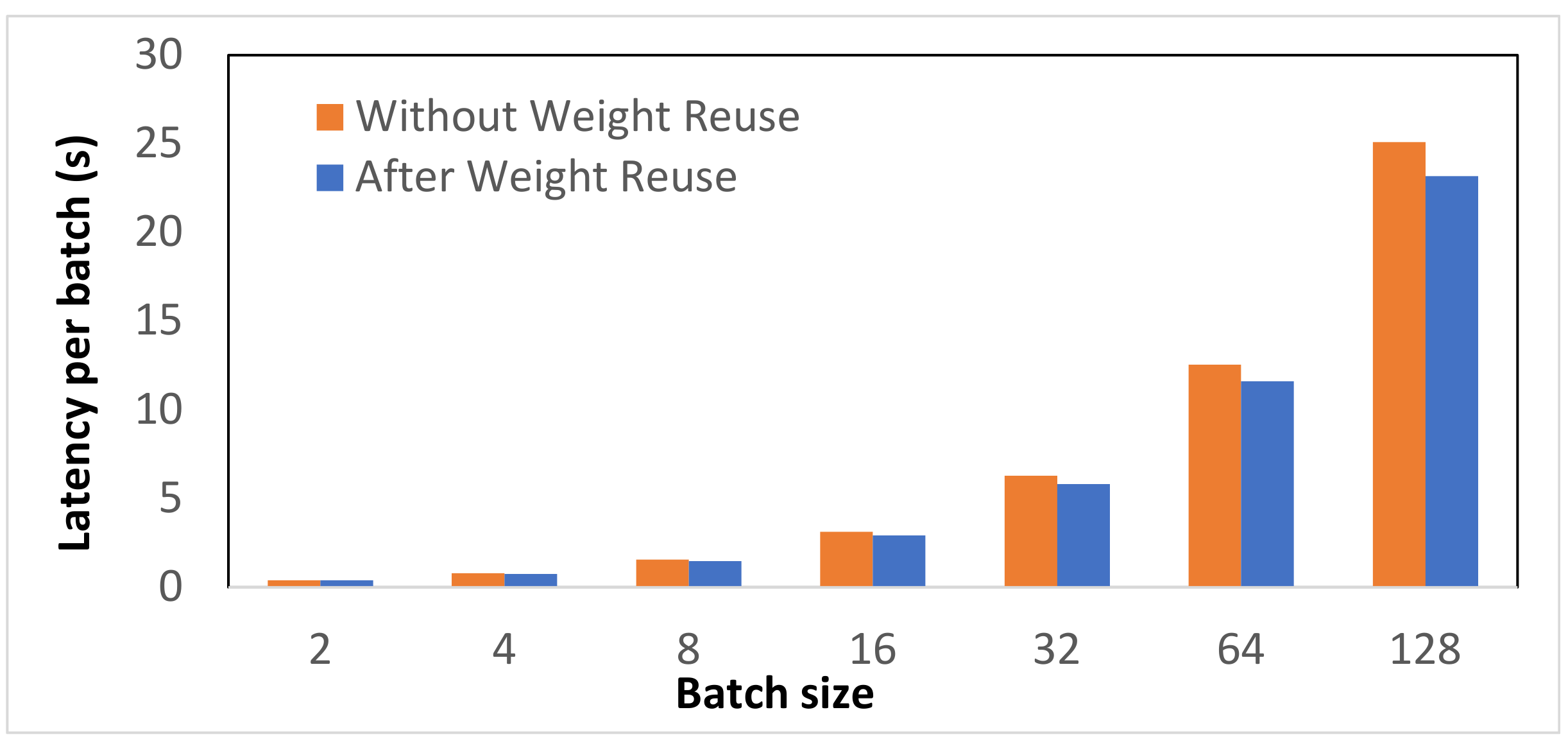}
  \caption{
  Experimental Results of The Data Reshaping Approach without Weight Reuse and after Weight Reuse}
  \label{fig:experiment_validate}
  \Description{fig:experiment}
\end{figure}

{
As mentioned in Section~\ref{sec:memory}, we optimize the DRAM access incrementally. We first achieve intra-tile continuous memory allocation by reorganizing the DRAM layouts for input features, output features, and weights. Then we re-schedule the loop order to achieve inter-tile continuous memory allocation. These two parts are combined to improve memory access continuity together when the batch size is 1. Considering the training process involves convolution operations among a mini-batch, we further propose and apply a weight reuse strategy based on the proposed data layout. Table~\ref{tab:effectiveness approach} shows the experimental results of the data reshaping approach without weight reuse and after weight reuse. The batch size is also 4, and the latency for FP/BP without and after weight reuse is nearly the same. It is because, in FP and BP, input features and weights are transmitted together. As can be seen from Section~\ref{sec:model}, When $\lceil\frac{Tn}{p}\rceil \times ((Tr^i-1)\times S^i+K^i)\times ((Tc^i-1)\times S^i+K^i)>\lceil\frac{Tm\times Tn}{p}\rceil\times K^i\times K^i$, $t^i_{IFM}>t^i_{WEI}$. Therefore, reusing the weights may not reduce the latency as a whole. However, in WU, the latency with weight reuse is apparently less than that without reuse. It is because the transmission of weights happens during storing the results in WU, which cannot be totally covered by $t^i_{COMP}$ (For example, in the last iteration of the loop in line 2 from Fig,~\ref{fig:inter} (b) and the loop in line 3 from Fig,~\ref{fig:inter} (c)). As a whole, reusing weights can reduce the latency of the training phase of the whole network. Fig.~\ref{fig:experiment_validate} shows the latency without and with weight reuse when the batch size ranges from 2 to 128. It shows that when the batch size increases, applying the reuse strategy has more apparent advantages than only achieving intra-tile and inter-tile data access continuity.}

\subsection{Accuracy of The Performance Model}
\label{sec:effectiveness model}
After displaying the effectiveness of the data reshaping approach, we use the AlexNet to validate the accuracy of the performance model.  Our scheduling tool first determines optimal tiling parameters which are shown in Table~\ref{tab:effectiveness model}. Then the latency is estimated by our model and tested on-board separately. As shown in Table~\ref{tab:effectiveness model}, the estimated results are close to the tested results. The results verify the accuracy of the performance model.
 
\begin{table}
  \caption{
  Experimental Results Validating The Performance Model}
  \label{tab:effectiveness model}
  \begin{tabular}{cccccc}
    \toprule
    AlexNet
    &Process
    &$[Tr^i, Tc^i, M^i\_on]$
    &Our Model (cycles)
    &On-board (cycles)
    &Deviation
    \\
    \midrule
    Conv 1
    &\multirowcell{3}{FP\\BP\\WU} 
    &\multirowcell{3}{$[2, 55, 96]$\\N/A\\$[2, 55, 96]$}
    &\multirowcell{3}{11,504,640\\N/A\\9,043,384}
    &\multirowcell{3}{11,419,835\\N/A\\9,299,086}
    &\multirowcell{3}{0.74\%\\N/A\\2.75\%}
    \\\\\\  \hline
    Conv 2
    &\multirowcell{3}{FP\\BP\\WU} &\multirowcell{3}{$[27, 27, 112]$\\$[27, 27, 48]$\\$[27, 27, 112]$}
    &\multirowcell{3}{7,309,808\\7,126,784\\7,423,616}
    &\multirowcell{3}{7,312,794\\7,146,578\\7,430,533}
    &\multirowcell{3}{0.04\%\\0.28\%\\0.09\%}
    \\\\\\  \hline
    Conv 3
    &\multirowcell{3}{FP\\BP\\WU} 
    &\multirowcell{3}{$[13, 13, 112]$\\$[13, 13, 112]$\\$[13, 13, 112]$}
    &\multirowcell{3}{2,478,272\\2,566,987\\2,682,240}
    &\multirowcell{3}{2,510,310\\2,671,392\\2,706,696}
    &\multirowcell{3}{1.28\%\\3.91\%\\0.90\%}
    \\\\\\  \hline
    Conv 4
    &\multirowcell{3}{FP\\BP\\WU} 
    &\multirowcell{3}{$[13, 13, 112]$\\$[13, 13, 112]$\\$[13, 13, 112]$}
    &\multirowcell{3}{3,646,400\\3,861,220\\3,960,960}
    &\multirowcell{3}{3,708,934\\3,972,757\\4,014,651}
    &\multirowcell{3}{1.69\%\\2.81\%\\1.34\%}
    \\\\\\  \hline
    Conv 5
    &\multirowcell{3}{FP\\BP\\WU} 
    &\multirowcell{3}{$[13, 13, 112]$\\$[13, 13, 112]$\\$[13, 13, 112]$}
    &\multirowcell{3}{2,432,368\\2,618,372\\2,640,640}
    &\multirowcell{3}{2,475,263\\2,686,910\\2,677,726}
    &\multirowcell{3}{1.73\%\\2.55\%\\1.38\%}
    \\\\\\  \hline
    \textbf{Total}
    & & &\textbf{69,295,691}
    &\textbf{70,033,465}
    &\textbf{1.05\%}
    \\
  \bottomrule
\end{tabular}
\end{table}

\subsection{CNN Training Performance}
\label{sec:performance}

In this section, we conduct end-to-end evaluations on different neural networks. We first compare our design with the automatic compiler-based FPGA accelerator~\cite{venkataramanaiah2019automatic}. {
It adopted a combination of channel-level parallelism and feature map-level parallelism with the unrolling factors for columns, rows, and output channels. It initially stored weights tile by tile in a transposable format in DRAM and read the entire weights of a Conv layer from DRAM to their on-chip buffer.} The baseline implemented a ‘1X’ CNN on the CIFAR-10 dataset with the structure as Conv 1 ($[M^i, N^i, R^i, C^i, K^i, S^i]=[16, 3, 32, 32, 3, 1]$) - Conv 2 ($[16, 16, 32, 32, 3, 1]$) - Pooling - Conv 3 ($[32, 16, 16, 16, 3, 1]$) - Conv 4 ($[32, 32, 16, 16, 3, 1]$) - Pooling - Conv 5 ($[64, 32, 8, 8, 3, 1]$) - Conv 6 ($[64, 64, 8, 8, 3, 1]$)- Pooling - FC ($[10, 1024, 1, 1, 1, 1]$), using 16-bit fixed-point precision. We test the same network on both PYNQ-Z1 and ZCU102 boards. The DMA stream bandwidth is 128 bits for ZCU102 and 32 bits for PYNQ-Z1. Our design focuses on implementing on-device FPGAs without sacrificing precision, so 32-bit floating-point is adopted. Vivado utilization report provides the utilization of BRAMs, DSPs, and the power report provides the total on-chip power. We measure the latency of training the whole batch with the batch size of 128. Then we calculate the latency per image and the throughput. 

\begin{figure}[h]
  \centering
  \includegraphics[width=4.8 in
  ]{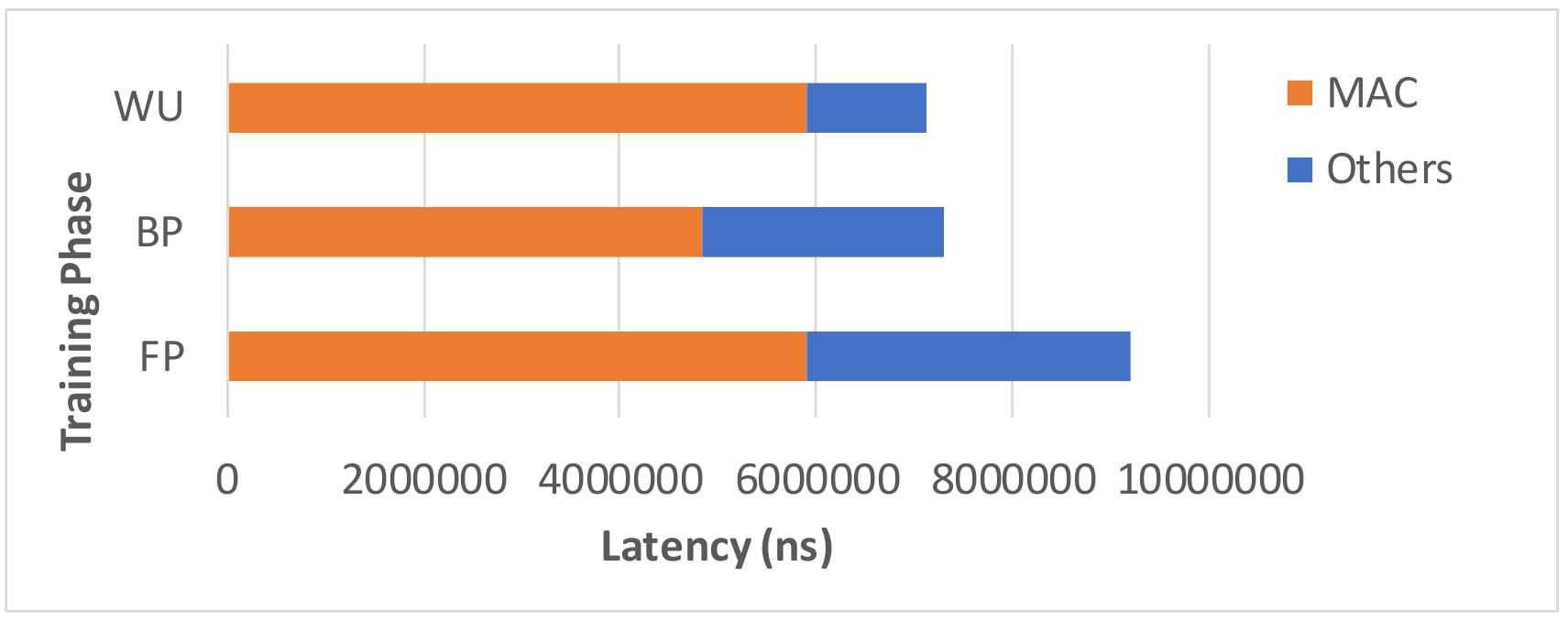}
  \caption{
  Latency breakdown of CIFAR-10 '1X' CNN for FP, BP
and WU when the batch size is 128.}
  \label{fig:experiment_1XCNN}
  \Description{fig:experiment}
\end{figure}

Table~\ref{tab:1XCNN} shows the comparison results between the baseline~\cite{venkataramanaiah2019automatic} and our design in terms of resource utilization, throughput, energy efficiency, etc. The Stratix 10 GX adopted in the baseline is an advanced FPGA board developed by Intel. It is unfair to compare the throughput directly for different devices. However, energy efficiency is an important metric to judge the performance of edge devices, thus we use energy efficiency as the metric for different designs on different FPGAs. We nominate the throughput and efficiency by multiplying the bit width of the data type. Although using the fixed-point data type is much more DSP-efficient and power-efficient than adopting floating-point under the same bit width, our nominal efficiency still can outperform that of the baseline. {
The reason is that the baseline has more data transmission latency especially for WU where accessing weight gradients, weights, and storing back the updated values leads to DRAM access latency. 51\% percent of the overall latency in one iteration of a batch is consumed in WU~\cite{venkataramanaiah2019automatic}. Fig.~\ref{fig:experiment_1XCNN} shows the latency breakdown of our design. The total latency for each training process is calculated by summarizing latency for each Conv layer, and the latency for MAC is the theoretical computation latency calculated by accumulating $t^i_{COMP}$ for each Conv layer based on the performance model. Since the '1X' CNN is a relatively small network, the number of loops is also small. According to the performance model, although double buffers are adopted, the computation and data transmission is conducted in sequential in the first and last iteration of the loop, while they are in parallel for the middle iterations. Therefore, when the number of loops is small, the proposed design also includes much data transmission latency for FP, BP, WU. However, other optimizations like loop order scheduling and weight reuse in a mini-batch reduce the number of off-chip memory access. Therefore, our computation latency is still much more than 50\% percent of the total latency in FP, BP, or WU, which takes up a larger proportion compared with the baseline.

Besides, the baseline stored the entire weights of a Conv layer from DRAM to the on-chip buffer. Their design cannot support denser networks where the on-chip buffer cannot hold the entire weights of each Conv layer. However, our design does not have such restrictions and can support many larger networks. 
} 

\begin{table}
  \caption{Experimental Results on the '1X' CNN}
  \label{tab:1XCNN}
  \begin{tabular}{cccc}
   \toprule
     
    & Baseline~\cite{venkataramanaiah2019automatic} 
    & Ours
    & Ours\\
    \midrule
    Platform
	 & Stratix 10 GX & PYNQ-Z1 & ZCU102\\ \hline
   Frequency (MHz)
    & 240 &100 &100\\ \hline 
  DSP Utilization
    & 1699 (30\%) & 212 (96.4\%) & 1315 (52.2\%)\\ \hline
 $D_{Conv}$ ($D_{Conv}/Used\ DSPs$) 
    &  & 180 (84.9\%) & 1280 (97.3\%)\\ \hline
  BRAM Utillization
    & 10.6 (4.4\%) & 123 (87.9\%) & 324 (35.5\%)\\ \hline
 $B_{Conv}$($B_{Conv}/Used\ BRAMs$) 
    &  & 108 (87.8\%) & 288 (88.9\%)\\ \hline
 Power (W)
    & 20.6 &1.85 (11.14X) &6.89 (2.99X)\\ \hline
 Data Type
	 & Fixed 16 &FP 32 &FP 32	\\ \hline
 Batch Size
    & 40 & 128 & 128\\ \hline
 Latency/Image (ms)
    &0.36 & 14.32 & 2.08\\ \hline
 Throughput 
    &163 GOPS & 4.08 GFLOPS & 28.15 GFLOPS\\ \hline 
 \multirowcell{2}{Nominal Throughput\\(GOPS$\times$ precision)}
   &\multirow{2}{*}{2608} &\multirow{2}{*}{130.56} &\multirow{2}{*}{900.8}\\\\ \hline
 Energy Efficiency
    &7.90 GOPS/W
    & 2.21 GFLOPS/W
    &4.09 GFLOPS/W\\ \hline
 \multirowcell{2}{Nominal Effciency \\(GOPS$\times$ precision/W)}
   &\multirow{3}{*}{126.4} &\multirow{3}{*}{70.72} &\multirow{3}{*}{130.88 (1.04X)}\\\\ 
  \bottomrule
\end{tabular}
\end{table}

In Table~\ref{tab:1XCNN}, the $D_{Conv}$ and $B_{Conv}$ are the DSPs and BRAMs numbers for the Conv layer estimated by our resource model. The percentage shows the ratio between the estimated resources and  actual resources used in the whole end-to-end training. As mentioned in Section~\ref{sec:resource} and Section~\ref{sec:schedule}, $D_{Conv}$ and $B_{Conv}$ are lower than the DSPs and BRAMs numbers used in the realistic end-to-end training process. For '1X' CNN, the extra on-chip resources mainly function for maximum pooling layers. Besides, a few DSPs are utilized to calculate BRAM addresses of features and weights.

\begin{figure}[h]
  \centering
  \includegraphics[width=5.4 in
  ]{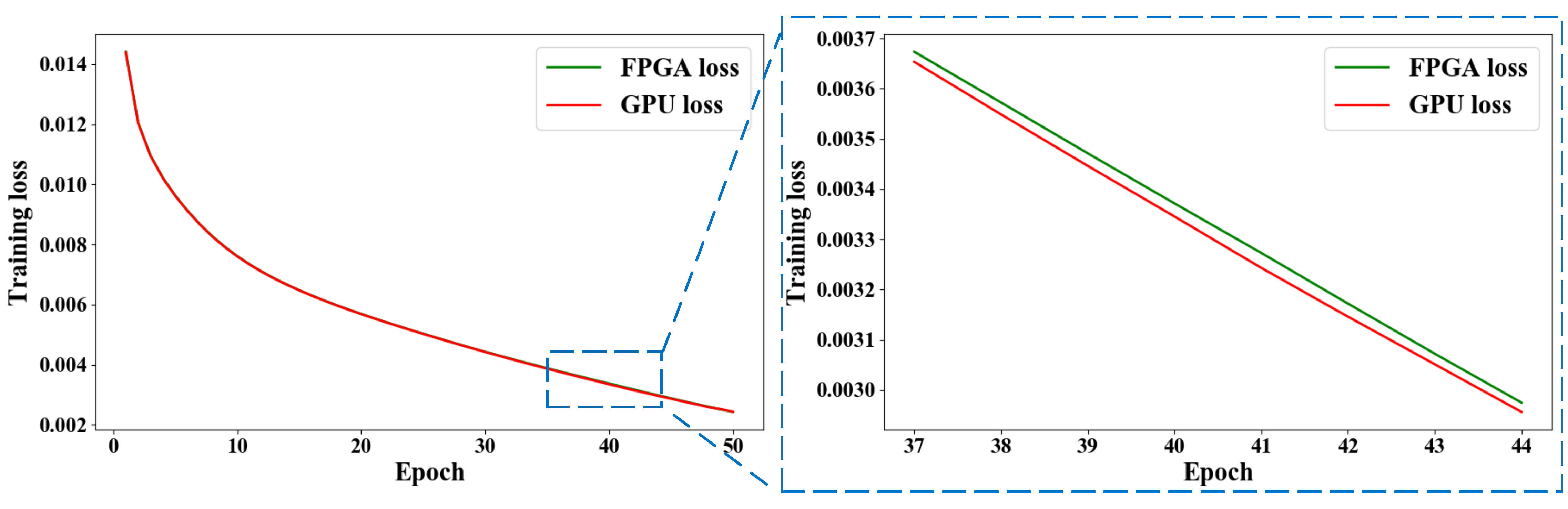}
  \caption{
  The loss curves during the training phase.}
  \label{fig:loss}
  \Description{fig:loss}
\end{figure}

{
To validate the correctness of our design, we also implement the whole training phase of the '1X' CNN on ZCU102 and compare the training result with that on GPU. We load the Cifar-10 dataset from the secure digital (SD) card to the DRAM and run 50 epochs. The batch size is also 128, and the learning rate is 0.008. We use the V-100 GPU from AWS to validate the training process. The loss curves are shown in Fig.~\ref{fig:loss}. Since we adopt full precision and have not changed the training algorithm, the training result should be nearly the same as that on GPU. As can be seen in Fig.~\ref{fig:loss}, the two curves are really close to each other. We also test the trained model on the test dataset. The test accuracy is 65.22\% running on GPU and 64.82\% running on FPGA.}

Most state-of-art works~\cite{venkataramanaiah2019automatic,luo2020towards} mainly implemented their design on Cifar-10 dataset whose input image is really small ($3\times 32\times 32$) compared to real-world on-device learning scenarios. To verify that our accelerator with the data reshaping approach can support larger networks with larger feature sizes, we test our design on AlexNet and Vgg-16 for ImageNet whose input image parameters are $3\times227\times227$ and $3\times224\times224$ respectively. 
\begin{figure}[h]
  \centering
  \includegraphics[width=4.8 in
  ]{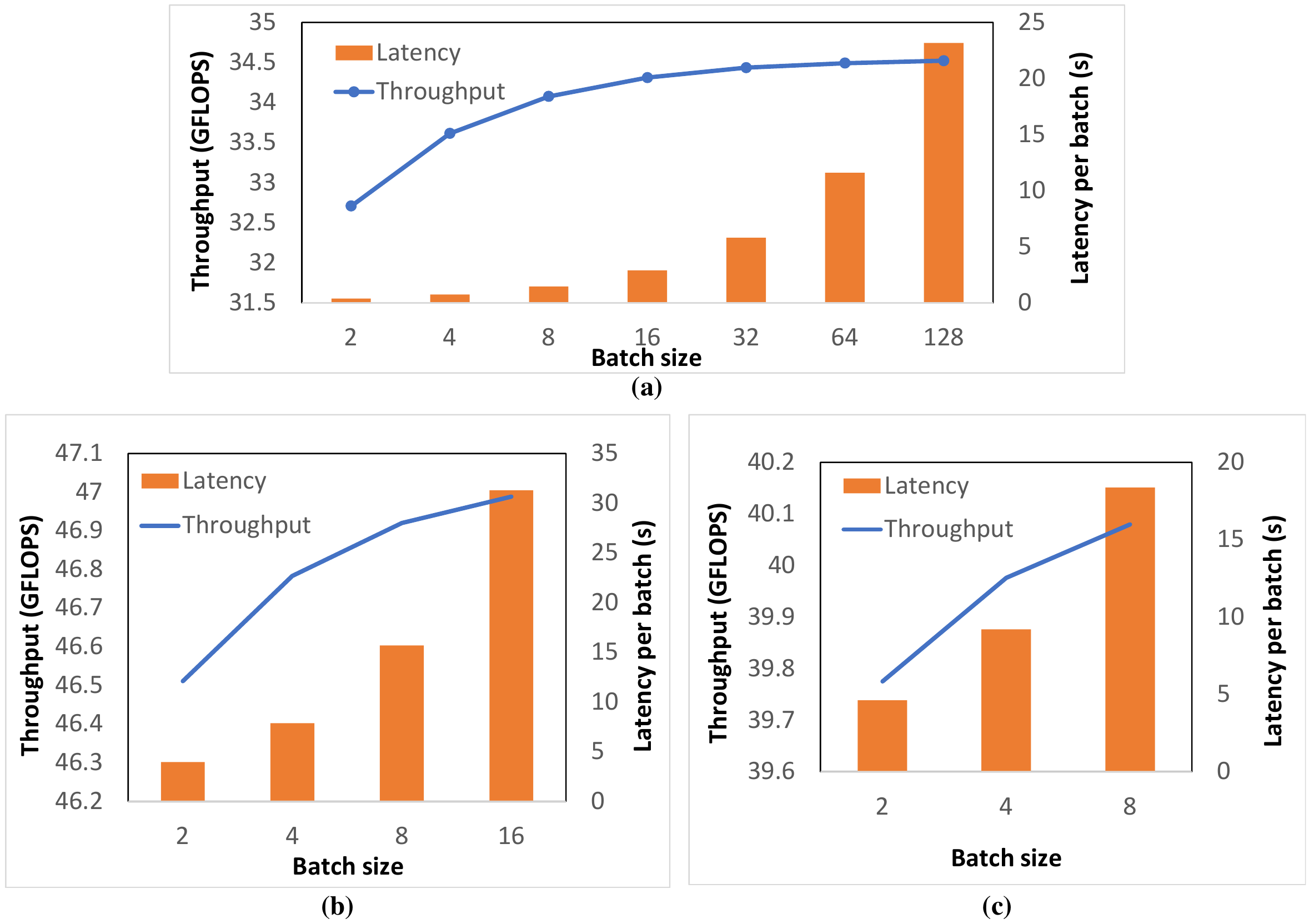}
  \caption{
  Experimental Results Training Different CNNs. (a) Throughput and Latency of AlexNet, (b) Throughput and Latency of Vgg-16 without BN layers, (c) Throughput and Latency of Vgg-16 with BN layers}
  \label{fig:experiment}
  \Description{fig:experiment}
\end{figure}

The data reshaping approach enables our accelerator to support end-to-end training with the following situations: (1) when the feature map size of a layer increases and the on-chip memory is not big enough to hold all the feature maps of the layer, and (2) when the number of channels increases and the weights buffer cannot hold all the weights of a layer. For AlexNet, its convolution kernel size ranges from 11×11 to 1×1 and feature map size ranges from 227×227 to 1×1, which covers the above-mentioned situations. {
Besides, the stride of the first Conv layer of AlexNet is 4. Implementing Conv layers with different stride sizes are more complex than only verifying the design on CNNs where the stride remains 1.} Therefore, AlexNet is ideal to verify that our design can support DNNs with a larger feature map size and larger weight density and can deal with different Conv layers shapes. Fig.~\ref{fig:experiment} (a) shows the throughput and latency of a batch for training the AlexNet model with batch size ranging from 2 to 128.  When the batch size is 128, the throughput reaches 34.52 GFLOPS. Because of weight reuse, the weights transmission bottleneck is ameliorated when the batch size increases, so the throughput in larger batch size is slightly higher than that for small batch size. However, unlike batch-level parallelism-based designs~\cite{luo2020towards} where the performance varies a lot under different batch sizes, the performance of our channel-level parallelism-based design is less affected by the batch size. As shown in Fig.~\ref{fig:experiment}, the throughput when the batch size is 2 is still above 32 GFLOPS.

We also test our design on Vgg-16 which has denser parameters, and the performance is shown in Fig.~\ref{fig:experiment} (b). Due to the DRAM memory size limitation of ZCU102, the maximum batch size is 16. As illustrated in Fig.~\ref{fig:experiment}, our design achieves higher throughput on Vgg-16 compared with AlexNet. It is because, for channel-level parallelism, the number of input channels of the first Conv layer is only 3, which is smaller than $Tn$, so the computation resources are not fully utilized in this layer. This effect is also mentioned in Section~\ref{sec:effectiveness approach}. However, such underutilization only happens in the first Conv layer and is alleviated when the neural network becomes deeper. Hence, in the deeper network, Vgg-16, we achieve higher throughput. 

{
To verify that our design can support the  BN layer which is a key component of typical CNN architectures, we also test the proposed design on Vgg-16 with BN layers. The performance is shown in Fig.~\ref{fig:experiment} (c). Apart from the loss and activation, the immediate BN parameters also need to be stored in DRAM. Due to the memory size limitation, the maximum batch size is 8. Unlike computation-intensive Conv layers, BN layers involve lots of data transmission processes. Some complex operations like extracting a root also cost extra computation resources and reduce the timing performance. Therefore, the overall throughput is a little less than that for Vgg-16 without BN layers.}

\begin{table}
  \caption{
  Experimental Results on AlexNet and Vgg-16}
  \label{tab:ImageNet}
  \begin{tabular}{cccc}
   \toprule
     
    Network 
    & AlexNet
    & Vgg-16 without BN
   & Vgg-16 with BN\\
    \midrule
  DSP Utilization
    & 1513 (60.0\%) 
    & 1508 (59.8\%)
    & 1680 (66.7\%)
    \\
  $D_{Conv}$ ($D_{Conv}/Used\ DSPs$) 
    & 1280 (84.6\%) 
    & 1280 (84.9\%)
    & 1280 (76.2\%)\\
  BRAM Utillization
    & 857 (94.0\%)
 & 787 (86.3\%)
 & 812 (89.0\%)\\
 $B_{Conv}$ ($B_{Conv}/Used\ BRAMs$)
    & 672 (78.4\%) 
    & 672 (85.4\%)
    & 672 (82.8\%)\\
 Power (W)
    & 7.736 & 7.712 & 8.208\\
 Batch Size
    & 128 & 16 & 8\\ 
 Throughput (GFLOPS)
    & 34.52 & 46.99 & 40.08\\ 
 Efficiency (GFLOPS/W)
    &4.46 & 6.09 & 4.88\\

  \bottomrule
\end{tabular}
\end{table}

Table~\ref{tab:ImageNet} also shows the resource utilization and energy efficiency of the FPGA for these networks. With the same estimated DSPs and BRAMs for Conv layers ($D_{Conv}$ and $B_{Conv}$), AlexNet requires more BRAMs than Vgg-16. It is because, compared to Vgg-16, AlexNet has a less regular weights kernel shape (ranging from $11\times 11$ to $1\times 1$), so we add an extra buffer to fetch a tile of weights from the on-chip Weight buffer before the Conv Kernel conducting MAC operations. Such optimization can release routing congestion caused by complex BRAM addresses calculation and allocation in FP, BP, and WU processes. Apart from the extra buffer, a small fraction of DSPs and BRAMs function for non-Conv layers. The accelerator also utilizes a few DSPs to calculate BRAM addresses. Therefore, as mentioned in Section~\ref{sec:schedule}, the estimated boundary of $D_{Conv}$ and $B_{Conv}$ in realistic end-to-end system design should be slightly smaller than the total DSPs and BRAMs numbers. From our experimental results, assigning $80\%$ of DSPs and $75\%$ BRAMs should be enough.

{
As for the Vgg-16 with BN layers, extra computation resources are utilized to do complex operations such as division, root extraction, etc. Therefore, Vgg-16 with BN layers costs more DSP resources compared with Vgg-16 without BN layers. Additional BRAMs are also utilized to buffer BN parameters for a batch.}

In our end-to-end training validation, we utilize 1508 DSPs for the Vgg-16 model. The theoretical peak performance with 1508 DSPs
on the 32-bit floating-point accelerator is $\frac{1508}{5}\times 2\times 0.1$ GHz$ = 60.3$ GFLOPS, while our attainable end-to-end test is 46.99  GFLOPS including pooling and ReLU operations.

\subsection{Comparison with State-of-art Works}
\label{sec:compare}
{
Comparisons of the best performance between our design and other state-of-art FPGA-based training accelerators are shown in Table~\ref{tab:compare}.} In the table, "N/A" means that the metric is not provided, and "$\approx$" means that the value is obtained by approximate estimation. Since the platforms, the neural networks for training, and the data type are different, it is extremely difficult to fairly compare between different training accelerators. However, our design still shows desirable performance even under such circumstances.

\begin{table*}
  \caption{
  Comparison of Different FPGA-based Training Accelerators}
  \label{tab:compare}
  \begin{tabular}{cccccc}
   \toprule
    Accelerator 
    &\multirowcell{2}{Chow et al.\\2017~\cite{liu2017fpga}}
    &\multirowcell{2}{DarkFPGA\\2020~\cite{luo2020towards}}
    &\multirowcell{2}{Seo et al.\\2020~\cite{venkataramanaiah2020fpga}}
    &\multirowcell{2}{FeCaffe\\2020~\cite{he2019fecaffe}}
    &Ours\\\\ 
    \midrule
  Platform
    & ZU19EG & XCVU9P & Stratix 10 MX & Stratix 10 & ZCU102\\ \hline
Technology
    & 16nm & 16nm & 14nm & 14nm & 16nm\\ \hline
    
 DSP Util.
    & 1500 & 4202  & 1040 & 1796 & 1508\\ \hline
 Freq. (MHz)
    & 200 & 200  & 185 & 253 & 100\\ \hline
 Power (W)
    & 14.24 & 13.5  &  $\approx$20 & N/A & 7.712\\ \hline
 Network
    & LeNet-10 & Vgg-like & ResNet-20 & AlexNet & Vgg-16\\ \hline
 Dataset
    & CIFAR-10 & CIFAR-10 & CIFAR-10 & ImageNet & ImageNet\\ \hline
 Data Type
	 & FP 32 & Fixed 8 & FP 16 & FP 32	& FP 32\\ \hline
 Throughput 
    &\multirowcell{2}{86.12\\GFLOPS} &\multirowcell{2}{1417\\GOPS} &\multirowcell{2}{$\approx$180\\GFLOPS}  &\multirowcell{2}{$\approx$24\\GFLOPS}
    &\multirowcell{2}{46.99\\GFLOPS}\\\\  \hline
 Energy Effi.
    &\multirowcell{2}{6.05\\GFLOPS/W} 
    &\multirowcell{2}{104.96\\GOPS/W}
    &\multirowcell{2}{$\approx$9\\GFLOPS/W} & N/A &\multirowcell{2}{6.09\\GFLOPS/W}\\\\ \hline
\multirowcell{3}{Nominal\\Thro.(GOPS\\$\times$ precision)}
   & 2755.84 & 11336 & $\approx$2880 & $\approx$768 & 1503.68\\\\\\ \hline
 \multirowcell{3}{Nominal\\Effi. (GOPS\\$\times$ precision/W)}
   & 193.6 & 839.68 & $\approx$144 & N/A & 194.88\\\\\\
  \bottomrule
\end{tabular}
\end{table*}


{
To better illustrate the uniqueness of the proposed design, we also compare our work with the accelerators that also adopted 32-bit floating-point. The comparisons are shown in Table~\ref{tab:chow} and Table~\ref{tab:fecaffe}. The design in~\cite{liu2017fpga} was tested on LeNet-10 which is a really small network with the structure as Conv 1 ($[M^i, N^i, R^i, C^i, K^i, S^i]=[32, 3, 32, 32, 3, 1]$) - Pooling - Conv 2 ($[32, 32, 16, 16, 3, 1]$) - Pooling - Conv 3 ($[64, 32, 8, 8, 3, 1]$) - Pooling - FC ($[64, 1024, 1, 1, 1, 1]$) - FC ($[10, 64, 1, 1, 1, 1]$). As explained in Section~\ref{sec:performance}, the underutilization of computation resources in the first Conv layer reduces the overall throughput. Therefore, the performance of the proposed design on this small network cannot be as superior as that in deeper networks like Vgg-16. However, our design is a general architecture that can support both small networks and larger networks, while the accelerator in~\cite{liu2017fpga} only targeted such small networks. It first achieved feature-map level parallelism in a uniform computation engine, and then unrolled channel-level parallelism factors to improve utilization of computation resources. For the memory access issues, the input and output features of each layer are all stored on the FPGA chip, which restricts the work from extending to support the networks where the on-chip BRAMs are not large enough to hold entire features of a Conv layer. However, larger networks like AlexNet and Vgg are commonly applied in practical applications. Unlike~\cite{liu2017fpga}, our work not only enables on-device training on larger CNN models but also achieves higher throughput when the network becomes deeper. Besides, the number of operations of LeNet-10 reported in~\cite{liu2017fpga} is 74.43 MFLOPs. However, according to $total\ number\ of\ training\ operations=2\times(3\times\sum_{i=1}^{n}{M^i\times N^i\times R^i\times C^i\times K^i\times K^i}-M^1\times N^1\times R^1\times C^1\times K^1\times K^1)$, the actual number of operations that we obtain is only 25.17 MFLOPs. In this formula, $2\times$ is due to the FP 32 data type, and $3\times$ is due to the fact that each layer needs to conduct FP, BP, and WU except the 1st layer which only needs to conduct FP and WU.    

The FeCaffe~\cite{he2019fecaffe} introduced a Caffe framework with OpenCL which can integrate FPGA to perform CNN network training. It only provided DSP utilization and throughput which are shown in Table~\ref{tab:fecaffe}. Compared to the FeCaffe framework, our design utilized fewer computation resources but achieved higher throughput implementing AlexNet.

}

\begin{table}
  \caption{
  Experimental Results on LeNet-10 Compared with Chow et al.~\cite{liu2017fpga}}
  \label{tab:chow}
  \begin{tabular}{ccc}
   \toprule
     
    & Chow et al.~\cite{liu2017fpga} 
    & Ours\\
    \midrule
    Platform
	 & ZU19EG  & ZCU102\\ \hline
   Frequency (MHz)
    & 200 &100 \\ \hline 
  DSP Utilization
    & 1699 (76.2\%) & 1315 (52.2\%)\\ \hline
  BRAM Utillization
    & 174 (17.7\%) &340 (37.3\%)\\ \hline
 Power (W)
    & 14.24  &7.14\\ \hline
 Throughput 
    &86.12 GFLOPS & 15.47 GFLOPS\\ \hline 
 Energy Efficiency
    & 6.05 GFLOPS/W
    &2.17 GFLOPS/W\\
 
  \bottomrule
\end{tabular}
\end{table}

\begin{table}
  \caption{
  Experimental Results on AlexNet Compared with FeCaffe~\cite{he2019fecaffe}}
  \label{tab:fecaffe}
  \begin{tabular}{ccc}
   \toprule
     
    & FeCaffe~\cite{he2019fecaffe} 
    & Ours\\
    \midrule
    Platform
	 & Stratix 10  & ZCU102\\ \hline
   Frequency (MHz)
    & 253 &100 \\ \hline 
  DSP Utilization
    & 1796 (31.2\%) & 1513 (60.0\%)\\ \hline
  BRAM Utillization
    & N/A &857 (94.0\%)\\ \hline
 Power (W)
    & N/A  &7.736\\ \hline
 Throughput 
    &$\approx$24 GFLOPS & 34.52 GFLOPS\\ \hline 
 Energy Efficiency
    & N/A
    &4.46 GFLOPS/W\\
 
  \bottomrule
\end{tabular}
\end{table}

{
The work in~\cite{venkataramanaiah2020fpga} also adopted both feature-map level parallelism and channel-level parallelism, similar to its preliminary work in~\cite{venkataramanaiah2019automatic}. The best nominal energy efficiency reaches 144 (GOPS $\times$ precision) which is lower than our best nominal energy efficiency which is 194.88 (GOPS $\times$ precision). As for the memory access issues, the accelerator in~\cite{venkataramanaiah2020fpga} targeted devices equipped with high bandwidth memory (HBM2). Compared with DMA, the HBM2 is superiorly advanced with 16 pseudo channels providing a high number of I/O
data pins. However, HBM2 is a new high-speed memory technology and is only integrated into a few modern FPGAs like Stratix 10 MX. Most FPGA-based edge devices still rely on DMA to communicate between the FPGA chip and off-chip DRAM. Besides,~\cite{venkataramanaiah2019automatic} and~\cite{venkataramanaiah2020fpga} only tested their designs on the Cifar-10 dataset where the input image size is only $32\times 32$ which is really small so that their on-chip BRAMs can easily hold $P$ entire feature maps, where $P$ is the unrolling factors in the channel dimension. However, our design can support both small and large feature map sizes.

DarkFPGA~\cite{luo2020towards} placed DRAM data layout in the channel-height-width-batch (CHWB) pattern based on its batch-level parallelism-based design.} It achieves higher nominal energy efficiency because the 8-bit fixed points can improve the energy efficiency and DSP efficiency out of proportion. The previous study has shown that if the data precision is no more than 8-bit, two MACs can be calculated on one Xilinx DSP48, reducing the DSP usage
by half~\cite{li2020edd}. However, for 32-bit floating-point, 1 MAC operation takes up 5 DSPs in the Xilinx FPGA board. Besides, XCVU9P is an extremely high-end cloud-level FPGA that has superior efficiency than commonly used edge FPGAs. However, as mentioned in Section~\ref{sec:Motivations}, the batch-level parallelism adopted by DarkFPGA only achieved high throughput when the batch size is large. From their experiments, when the batch size is below 16, its throughput is below 100 GOPS which is around 800 GOPS$\times$precision after nominating, while our nominal throughput is stably above 1000 GOPS$\times$precision among different batch sizes. {
Besides, same with~\cite{venkataramanaiah2019automatic,venkataramanaiah2020fpga}, DarkFPGA also implemented their design on Cifar-10 dataset with a relatively small feature map size.}

\section{Conclusion}
\label{sec:conclusion}

In this paper, we design EF-train, an efficient DNN training accelerator enabling edge FPGAs to continuously learn on the device, which makes it possible for current FPGA-based edge-level applications to achieve domain adaption and personalization. We propose an FPGA-based CNN training accelerator with a unified convolution kernel to process FP, BP, and WU with full precision and a data reshaping approach to ensure continuous memory access during end-to-end training processes. We implement end-to-end CNN training effectively for low-power edge devices with restricted resources.  The experimental results show that our design achieves 46.99 GFLOPS and 6.09 GFLOPS/W in terms of throughput and energy efficiency, respectively.

\bibliographystyle{unsrt}
\bibliography{sample-base}










\end{document}